\documentclass[pdflatex]{sn-jnl}
\usepackage[numbers]{natbib}
\usepackage{graphicx}\usepackage{multirow}\usepackage{amsmath,amssymb,amsfonts}\usepackage{amsthm}\usepackage{mathrsfs}\usepackage[title]{appendix}\usepackage{xcolor}\usepackage{textcomp}\usepackage{manyfoot}\usepackage{booktabs}\usepackage{algorithm}\usepackage{algorithmicx}\usepackage{algpseudocode}\usepackage{listings}\usepackage{tabularx}
\usepackage{longtable}
\usepackage{array}
\usepackage{floatpag}
\usepackage{pdflscape}
\floatpagestyle{empty}

\theoremstyle{thmstyleone}

\theoremstyle{thmstyletwo}

\theoremstyle{thmstylethree}

\begin{document}

\title[Evaluation on Real Point-of-Care Clinical Queries]{
\centering
Expert Evaluation of Clinical AI Tools on Real Point-of-Care Clinical Queries
}

\author*[1]{\fnm{Jean} \sur{Feng}}\email{jean.feng@ucsf.edu}

\author[2, 3]{\fnm{Vishal} \sur{Patel}}

\author[4]{\fnm{Patrick} \sur{Heagerty}}

\author[5]{\fnm{Yifan} \sur{Mai}}

\author[1]{\fnm{Venkatesh} \sur{Sivaraman}}

\author[1]{\fnm{Patrick} \sur{Vossler}}

\author[1]{\fnm{Jialin} \sur{Ouyang}}

\author[2, 6, 7]{\fnm{Anupam B.} \sur{Jena}}

\affil*[1]{\orgdiv{Department of Epidemiology and Biostatistics}, \orgname{University of California, San Francisco}, \orgaddress{\city{San Francisco}, \state{California}, \country{USA}}}

\affil[2]{\orgname{Harvard Medical School}, \orgaddress{\city{Boston}, \state{Massachusetts}, \country{USA}}}

\affil[3]{\orgname{Brigham and Women's Hospital}, \orgaddress{\city{Boston}, \state{Massachusetts}, \country{USA}}}

\affil[4]{\orgdiv{Department of Biostatistics}, \orgname{University of Washington}, \orgaddress{\city{Seattle}, \state{Washington}, \country{USA}}}

\affil[5]{\orgdiv{Center for Research on Foundation Models (CRFM)}, \orgname{Stanford University}, \orgaddress{\city{Stanford}, \state{California}, \country{USA}}}

\affil[6]{\orgname{Massachusetts General Hospital}, \orgaddress{\city{Boston}, \state{Massachusetts}, \country{USA}}}

\affil[7]{\orgname{National Bureau of Economic Research}, \orgaddress{\city{Cambridge}, \state{Massachusetts}, \country{USA}}}

\abstract{
Physicians now pose millions of clinical questions to AI tools each week, yet these tools are evaluated largely on hypothetical or exam-style questions, not those actually asked in practice.
We report a blinded evaluation built on 620 Real-world Point-Of-Care Queries (Real-POCQi) submitted to the OpenEvidence (OE) platform by physicians spanning 30 specialties, as well as 187 questions from HealthBench.
149 practicing physicians across 36 states made head-to-head comparisons between answers from three frontier general-purpose models (Claude Opus 4.8, Gemini 3.1 Pro, and GPT-5.5) and a specialized clinical tool (OE), with graders matched to each question's specialty.
When comparing answers along five dimensions relevant to clinical decision support---accuracy, clinical utility, source quality, verifiability, and completeness---physicians scored the specialized tool highest on all axes; in the primary analysis on Real-POCQi, win differences (margins between win and loss rates) ranged from 25 to 39 percentage points (\textit{p}$<$0.001).
Results remained consistent in sensitivity analyses stratifying by citation display, answer length, OE-user status, and Real-POCQi versus HealthBench.
In parallel, LLM judges were found to systematically differ from expert judges, though both generally agreed on the best model.
These findings underscore two conclusions.
First, AI tool evaluations should reflect real-world query distributions and use expert judges that mirror the specialization defining modern medicine.
Second, the consistent advantage of the specialized tool over general-purpose models does not mean that the latter cannot serve similar purposes, but that targeted engineering and customization can yield meaningful gains in performance and utility for its users.
We release Real-POCQi as a public benchmark for further analysis, as well as the prespecified statistical analysis for reproducing results of this study.
}

\maketitle

\section{Introduction}\label{sec:intro}

Clinicians are among the rapid adopters of AI tools, with tens of millions of clinical requests made each month by U.S. physicians \citep{OffcallUnknown-qz}. Given the stakes of the decisions these tools inform, and the speed at which their capabilities and adoption are expanding, rigorous and safety-focused evaluation is essential.  This imperative has spurred a proliferation of research evaluating medical AI models \citep{Jin2021-fn, Arora2025-fp, Bedi2026-sx, Vishwanath2026-io}, but building evaluations that faithfully reflect real clinical practice remains a fundamental challenge \citep{Alaa2025-pq}.

Medical AI evaluation has largely relied on standardized benchmarks — fixed, shared datasets of test questions with predetermined correct answers or grading rubrics, such as those drawn from licensing examinations.
There is growing concern that these do not reflect the real-world challenges physicians face when using these models in practice, due to fundamental limitations in both how evaluation data are collected and how answers are graded.
First, public datasets rarely capture the real-world distribution of questions physicians ask during care delivery, in part because such queries are difficult to collect at scale and are seldom released by the proprietary platforms through which they are posed.
Early benchmarks relied heavily on multiple-choice items drawn from standardized licensing examinations \citep{Jin2021-fn, Singhal2023-sc}, while more recent efforts use questions co-designed by physicians and AI agents \citep{Arora2025-fp}.
In both cases, the questions are hypothetical by default, tend to oversimplify the clinical situation, and reflect speculative rather than empirically grounded assumptions regarding questions asked at the point of care.
Second, due to the difficulty in obtaining high-quality annotations in medicine, grading of AI on clinical topics typically rely on a small number of physicians to rate questions across many domains, disregarding the specialization that defines modern medicine.
Furthermore, fixed rubrics are commonly used, on the supposition that this grading scheme is more cost-efficient \citep{Vishwanath2026-io}.
However, recent works have shown that rubrics are substantially less accurate than conducting head-to-head comparisons, particularly in high-expertise domains \citep{Yang2026-pr}, because rubrics not only presuppose what makes an answer good but also make blanket assumptions.
This does not reflect the real-world needs of practicing physicians at the point of care. 
At the same time, there is now a debate regarding whether one can forgo human experts altogether by using a LLM judge or jury instead \citep{Chehbouni2025-ao, Bedi2026-sx, Vossler2026-ii}.
For AI-based clinical decision support systems, the key question is whether language models can reliably arbitrate clinical reasoning for questions that arise at the point-of-care, which has been largely untested up to now.
Given these gaps, there is a need to move beyond standardized benchmarks toward evaluation grounded in the realities of clinical practice, built on the questions clinicians ask and the judgment of appropriately matched experts.

\begin{figure}[tbp]
  \centering
  \includegraphics[width=\linewidth]{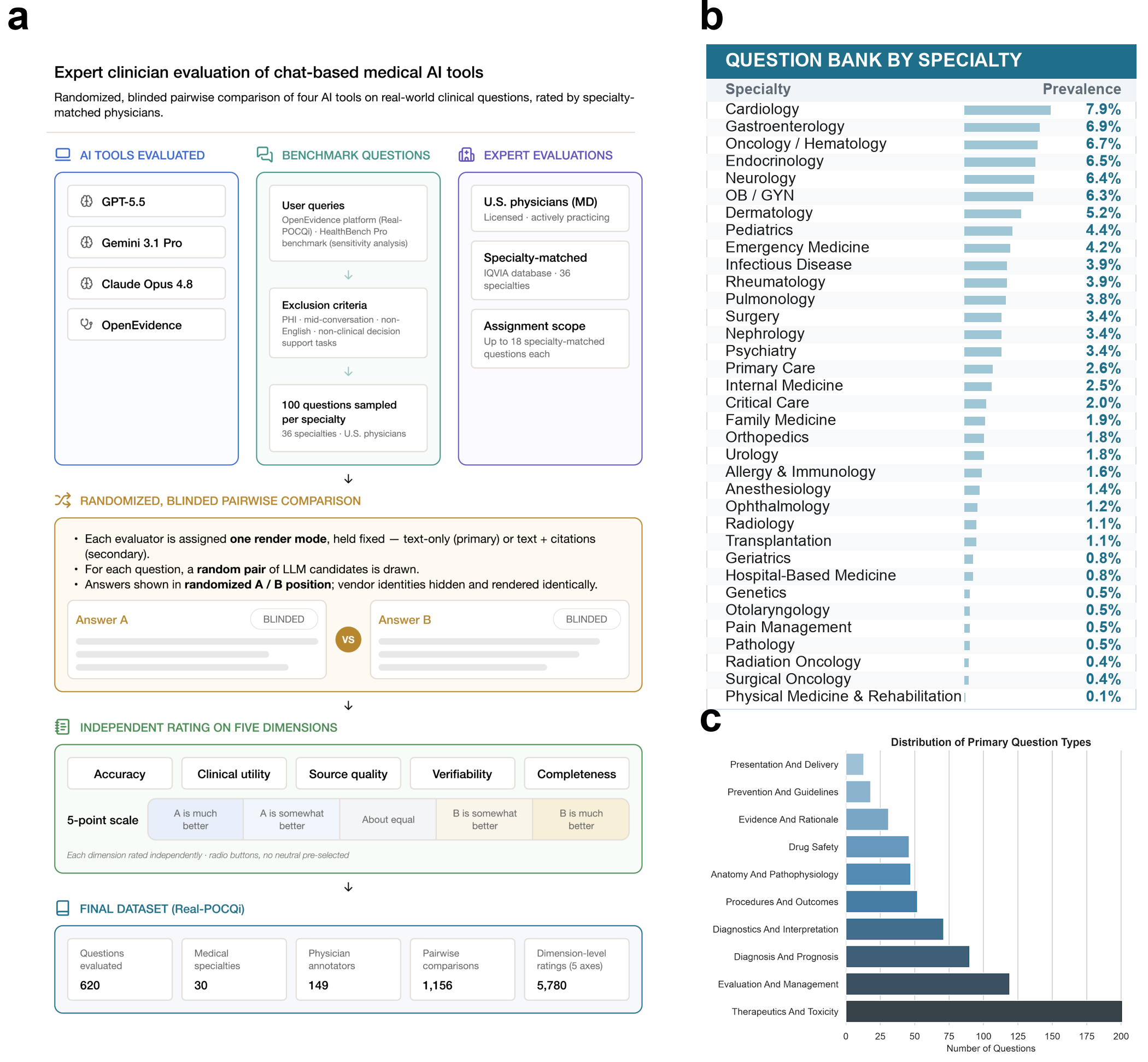}
  \caption{\textbf{Study design and composition of the benchmark dataset.} Specialty-matched, licensed and actively practicing U.S. physicians evaluated four chat-based medical AI systems --- GPT-5.5, Gemini 3.1 Pro, Claude Opus 4.8 and OpenEvidence (OE) --- on real-world clinical decision support questions in a randomized, blinded, pairwise design. \textbf{a}, Study schematic.
  The question bank (Real-POCQi) was derived from deidentified and rewritten OE real user queries, sampled across 36 specialties (IQVIA database) and filtered per exclusion criteria; HealthBench questions were additionally included for sensitivity analysis.
  Each evaluator was assigned a single render mode held fixed throughout --- text-only (primary) or text with citations (secondary); for every question a random pair of the four systems was drawn, and the two answers were shown in randomized A/B positions with vendor identities hidden and rendered identically. Each comparison was rated independently on five dimensions (accuracy, clinical utility, source quality, verifiability and completeness) on a five-point preference scale from A much better to B much better. \textbf{b}, Prevalence of each specialty area among sampled and rated questions in the dataset. \textbf{c}, Distribution of question types across the dataset, as identified through LLM-based clustering.}
  \label{fig:study_design}
\end{figure}

Here we report a large, blinded evaluation built on Real-world Point-Of-Care Queries (Real-POCQi) submitted by physicians to the OpenEvidence (OE) platform, with a sensitivity analysis on questions from the HealthBench benchmark \citep{Arora2025-fp} (Figure~\ref{fig:study_design}).
We publicly release the questions, answers, ratings, and analysis code for Real-POCQi (\url{https://huggingface.co/datasets/jjfenglab/Real-POCQi}, \url{https://github.com/jjfenglab/Real-POCQi-statistics}).
Questions in Real-POCQi were drawn from the more than one million questions submitted to OE each day by physicians across the U.S., spanning topics from diagnosis and prognosis to treatment selection and drug safety, and specialties from generalist domains such as primary care to subspecialties such as nephrology.
In a survey sent to randomly selected physicians across the U.S., 149 respondents across 36 states were matched to questions by clinical topic and made blinded, head-to-head comparisons between answers from four AI models: three frontier general-purpose models (Claude Opus 4.8, Gemini 3.1 Pro, and GPT-5.5) and one specialized clinical tool (OE).
Through the pairwise comparison approach popularized by LLM arenas \citep{Chiang2024-tw}, answers by AI systems were assessed along five dimensions related to point-of-care clinical decisions (accuracy, clinical utility, source quality, verifiability, and completeness). The final Real-POCQi dataset constitutes the largest specialty-matched public benchmark of real-world clinical queries asked by practicing clinicians, with 620 questions, 1156 comparisons, and 30 specialties.
With it, we evaluate the ability of general-purpose and specialized AI models to generate answers that reflect the needs of different medical domains.

\section{Results}\label{sec:results}

\subsection{Overview of the benchmark and evaluation study}
Based on a data collection plan agreed upon by the authors and OE, OE generated and emailed surveys to physicians sampled from the IQVIA OneKey provider database.
Survey questions were drawn from a sample of 3600 questions from the OE platform submitted by users across 36 specialties and, for the purposes of a sensitivity analysis, questions from HealthBench \citep{Arora2025-fp}.
The final Real-POCQi benchmark, composed of questions with at least one rating, includes 620 questions from physicians spanning 30 medical specialties (Figure~\ref{fig:study_design}a).
The top specialties represented in benchmark questions were cardiology, gastroenterology, and hematology/oncology (Figure~\ref{fig:study_design}b).
LLM-based clustering \citep{Grootendorst2022-it, Feng2026-ka} of the questions uncovered eleven key themes, with ``Therapeutics and Toxicity'' as the most prevalent, followed by ``Evaluation and Management'' and then ``Diagnosis and Prognosis.'' (Figure~\ref{fig:study_design}c).
Sample questions are provided in Table~\ref{tab:sample_questions} of the Appendix.

Answers to questions in the benchmark dataset were graded by 149 practicing U.S. physicians, generating 1156 pairwise ratings on five axes (Accuracy, Clinical utility, Source quality, Verifiability, and Completeness) between the four blinded AI models.
Accuracy was chosen as it is paramount in any setting that responses be accurate, regardless of writing style.
Clinical utility, source quality, and verifiability were chosen as these dimensions are uniquely relevant to clinical decision support: the ability to practice on the basis of the information provided and the ability to trust the information provided are correlated but independent from pure accuracy. Completeness is a complementary measure to evaluate when a response may be technically correct but incomplete.
Questions were randomly sampled so that the grader's specialty matched the clinical topic of the query.

To decouple the effects of how AI responses are rendered, a 2:1 random split was used to assign physicians to view only the text of the AI-generated responses (text-only) or both the text and citations (text+citation), respectively.
Median submission time was 127 seconds; ratings with submission times under 10 seconds were excluded from primary analyses (n = 18 excluded).
The inter-rater agreement rate---evaluated between physicians rating the same question and AI model pair within the same rendering mode---ranged from 74.1\% to 76.9\% across the five evaluation dimensions.
Weighted Cohen's kappa with quadratic weights was moderate to high, ranging from 23\% to 38\% for all but the verifiability axis which had a kappa of 9\%.
Users and non-users of OE were sampled evenly, resulting in 52.3\% of graders having registered accounts with OE.
Full descriptive statistics are provided in Appendix Table~\ref{tab:descriptives}.

\begin{figure}[tbp]
  \centering
  \includegraphics[width=0.8\linewidth]{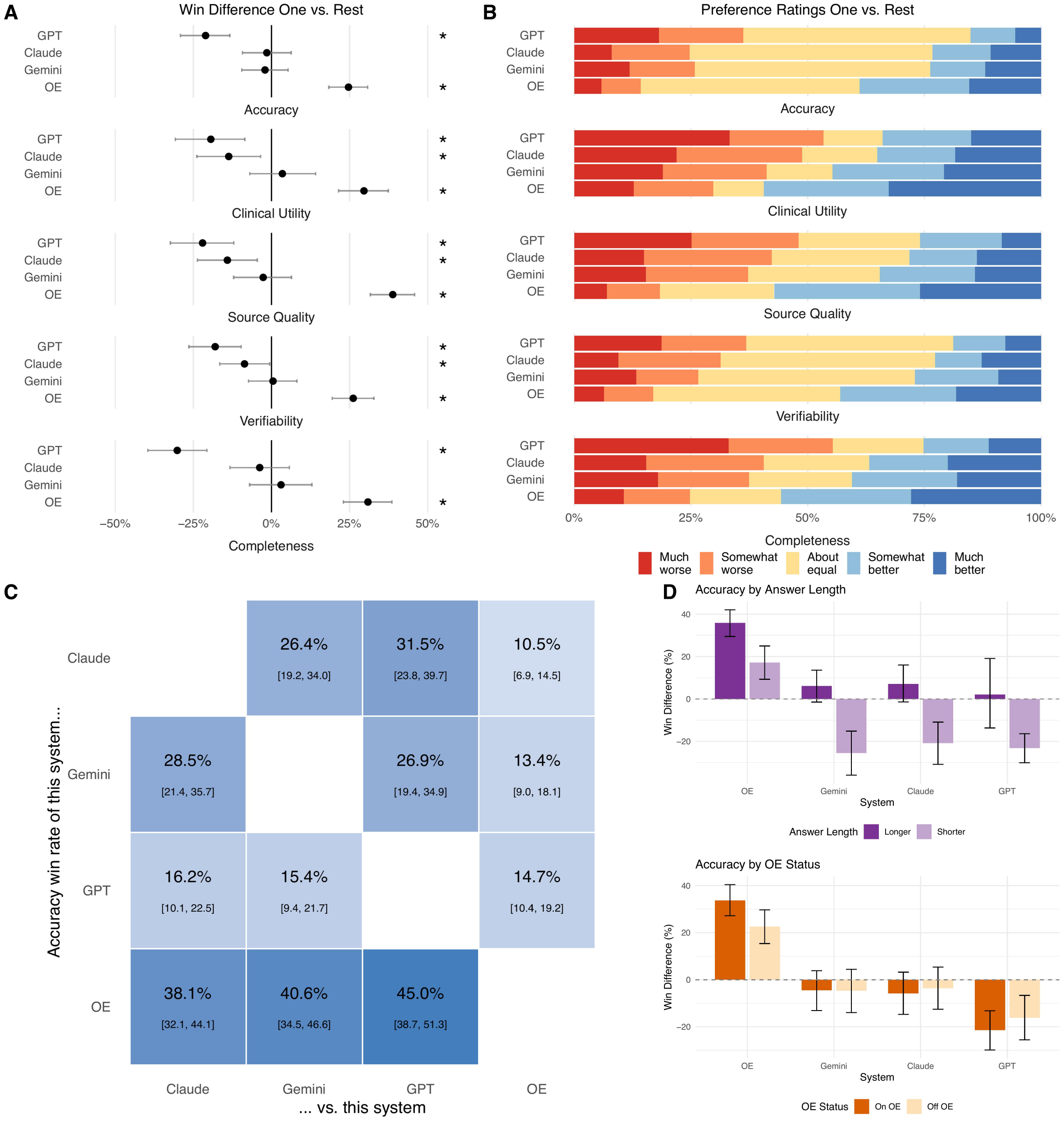}
  \caption{
  \textbf{Win differences and win rates of AI models in answering real-world clinical decision support queries.} All panels summarize blinded comparison of text-only answers, where physicians graded responses to questions in their clinical specialty.
  Answers were compared on a 5-point Likert scale, allowing for ties.
  For systems A and B, the win rate of A compared to B is how often A was rated as being better than B.
  The primary outcome, the One-vs-Rest (OvR) win difference, is the average difference between the win and loss rates when system A is compared to a comparator system.
  \textbf{a}, OvR win differences for GPT-5.5, Gemini 3.1 Pro, Claude Opus 4.8 and OpenEvidence (OE) across all five evaluation dimensions.
  \textbf{b}, The average distribution of preference ratings of a system compared to other comparators.
  \textbf{c}, Pair-wise win rates $P(\mathrm{row} > \mathrm{column})$ for every ordered pair of systems, where darker cells denote higher win rates.
  \textbf{d}, Sensitivity analyses of win differences along the central axis of accuracy, stratifying the primary analysis by whether the answer was longer versus shorter than its comparator (top) and stratifying by whether the evaluating clinician is an existing OE user (bottom).
  Stars indicate statistical significance at the level of $p=0.05$.
  All error bars are 95\% bootstrap confidence intervals.
  }
  \label{fig:main_summary}
\end{figure}

\subsection{Primary endpoint: Win differences between AI systems}

The prespecified primary endpoint of the study was a one-vs-rest win difference, defined as $(Wins - Losses) / N$, where the number of wins was how often a system was preferred to another, the number of losses was how often it was disfavored, and $N$ is the total number of comparisons for that system \citep{Pocock2012-md, Chiang2024-tw, Fang2026-cv}.
This metric ranges from $-100\%$ (always disfavored) to $100\%$ (always preferred), with zero indicating balanced wins and losses.

The primary analysis tested whether the win difference was zero for a given model and evaluation axis, when evaluators were only presented text-only answers to queries from Real-POCQi.
Only OE achieved positive win differences for all five evaluation axes, with two-sided p-values all $<$0.001 (summary in Figure~\ref{fig:main_summary}, details in Table~\ref{tab:p_values} of the Appendix).
For the central accuracy score, OE attained a win difference of 24.7\% (18.4 to 30.8\%, $p<$0.001), compared with $-21.1\%$ ($-29.1$ to $-13.3\%$, $p < 0.001$) for GPT-5.5, $-1.5\%$ ($-9.3$ to $6.3\%$, $p = 0.70$) for Claude Opus 4.8, and $-2.1\%$ ($-9.5$ to $5.3\%$, $p = 0.59$) for Gemini 3.1 Pro.
For clinical utility, OE achieved 29.6\% (21.5 to 37.4\%, $p < 0.001$), versus $-19.4\%$ for GPT-5.5, $-13.7\%$ for Claude Opus 4.8, and 3.5\% for Gemini 3.1 Pro.
Comparisons were directionally aligned for the other axes.
Of note, the highest win difference was 38.8\% (31.7 to 45.8\%, $p < 0.001$), achieved by OE along the dimension of source quality.
GPT-5.5 had the lowest win differences for all five evaluation axes (all $p<0.001$).

A comprehensive set of sensitivity analyses were conducted to assess the robustness of these findings.
First, win rates were recalculated for the setting where both text \textit{and} citations were displayed.
Results were qualitatively similar, where the main difference was that GPT-5.5 and Gemini 3.1 Pro had the lowest win differences for accuracy and Claude Opus 4.8 had lowest win difference for clinical utility (Appendix Figure~\ref{fig:winrate_by_model_textcitations}).
Second, a length-stratified analysis examined whether the win difference varied based on answer length (Figure~\ref{fig:length_stratified_winrate_by_model}), to assess whether the length of answers was associated with physician ratings.
On average, the length of answers from Claude Opus 4.8, Gemini 3.1 Pro, and OE were similar (Figure~\ref{fig:answer_length_distribution}), where pair-wise comparisons of their answer lengths showed that the probability of one system having a longer answer ranged from 42.8\%–57.2\%; in pairwise comparisons, answers from GPT-5.5 were the shorter of the two at a rate of 83.4\%.
In general, ratings for accuracy and completeness were most sensitive to answer length, as win differences for all AI models became positive when their answers were longer.
Ratings for source quality were the least sensitive to answer length, as the signs of the win differences were the same regardless of relative answer lengths.
The effects on clinical utility and verifiability were mixed, with only some win differences switching sign.
Third, when the same analysis was conducted on the sampled HealthBench questions, the same trends emerged.
OE was the only AI system with positive win differences across all five axes, though the effects were attenuated.
On HealthBench, OE's accuracy win difference was 14.1\% ($p=0.01$) and all other axes ranged from 18.8-27.6\% with $p<0.01$.

Given the widespread use of OE in clinical practice, we conducted a sensitivity analysis that stratified physician graders by whether their email was registered with an OE account.
Analyzing by OE user status is necessary to assess the potential bias during evaluations, which can remain despite the blinding strategies used; for instance, OE users may be habituated to OE responses or non-OE users may not use the tool specifically because they find it challenging to assess the information.
Stratification of the win rates analysis on the accuracy axis between the OE-user and non-OE-user subgroups is shown in Figure~\ref{fig:main_summary}d.
The results remained directionally identical for other axes (Table~\ref{tab:p_values}), with OE scoring positively on the other four axes in both subgroups ($p<0.001$), though the effect was attenuated among non-OE-users.

\subsection{Secondary endpoint: Win-rate matrices between AI systems}

The secondary endpoint was pairwise win-rate matrices, defined as the proportion of head-to-head comparisons in which system $A$ was preferred over system $B$.
Win-rate matrices provide a more granular characterization of relative performance than the one-vs-rest win difference (Figure~\ref{fig:main_summary}b, Appendix Figure~\ref{fig:appendix_pairwise_matrices}).

For accuracy, the AI systems formed three tiers (Figure~\ref{fig:main_summary}b).
GPT-5.5 had the lowest win-rates of 15.4\% (9.4--21.5\%), 16.2\% (10.0--22.7\%), and 14.7\% (10.4--19.2\%) in comparisons against Gemini 3.1 Pro, Claude Opus 4.8, and OE, respectively.
Gemini 3.1 Pro and Claude Opus 4.8 performed at near parity with one another, with Gemini preferred over Claude 28.5\% (21.3--35.9\%) of the time and Claude over Gemini 26.4\% (19.3--34.2\%) of the time.
OE outperformed the three general-purpose AI systems, with accuracy win rates of 45.0\% (38.8--51.3\%) against GPT-5.5, 40.6\% (34.6--46.9\%) against Gemini 3.1 Pro, and 38.1\% (32.1--44.3\%) against Claude Opus 4.8.
Similar patterns were observed for each of the other axes, with OE outperforming the other systems (Appendix Figure~\ref{fig:appendix_pairwise_matrices}).
Of note, general-purpose AI systems tended to compare most similarly in terms of their verifiability win rates.

As a sensitivity analysis, strict win-rate matrices were computed in which a system was credited with a win only when it was \textit{strongly} preferred (Appendix Figure~\ref{fig:appendix_pairwise_matrices}).
Under this more stringent criterion, the magnitude of all the win rates decreased as expected, but they followed the same patterns and results remained statistically significant.

\subsection{Evaluation by LLM-as-a-judge versus clinical experts}

\begin{figure}[tbp]
  \centering
  \includegraphics[width=0.9\linewidth]{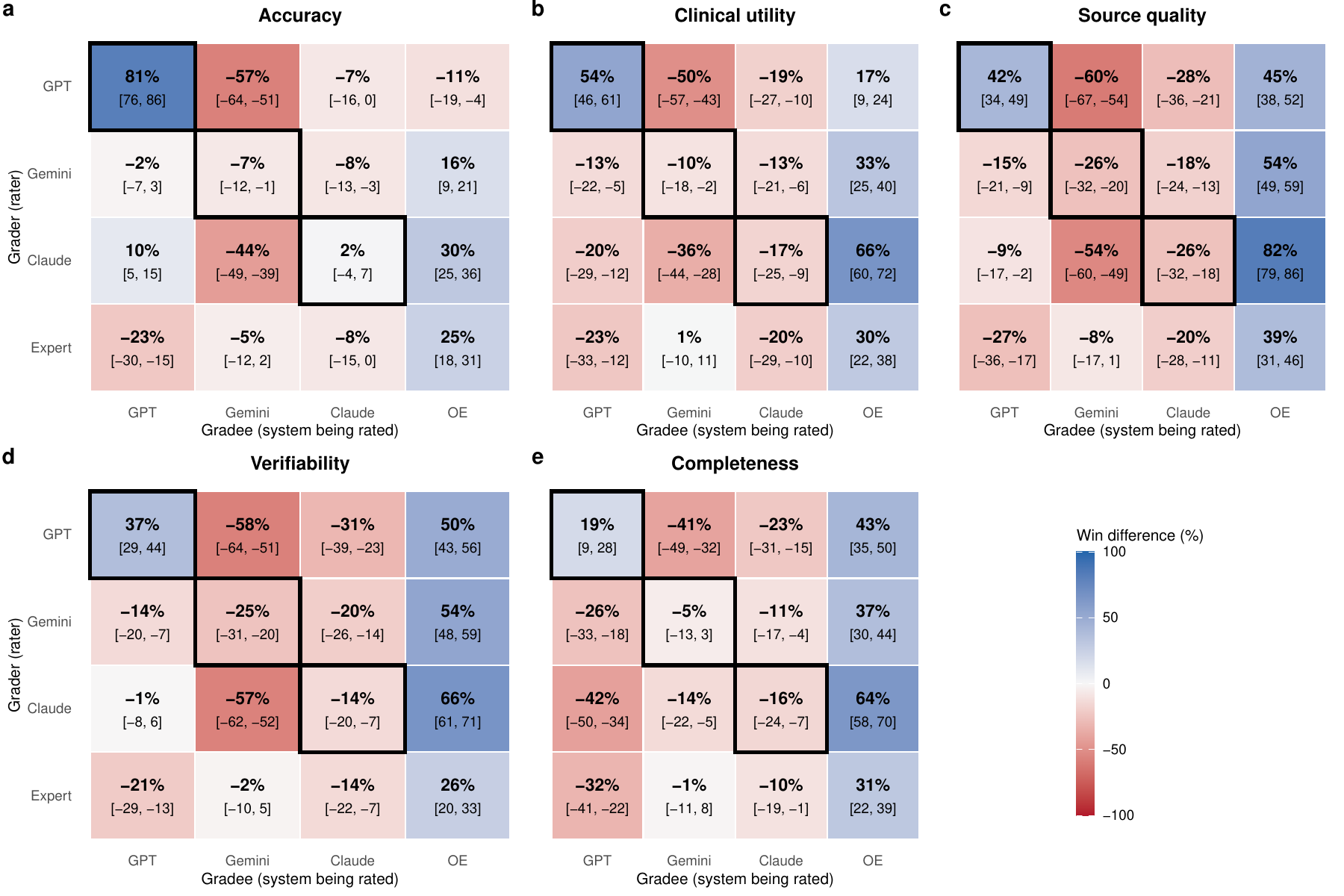}
  \caption{
  \textbf{AI system rankings from LLM-as-a-judge systems versus human experts.} Each panel is a grader $\times$ gradee heatmap of the one-vs-rest win difference for one evaluation axis.
  For a given grader (row), each cell is the win difference assigned to the gradee (column): its percentage of wins minus losses across all head-to-head comparisons against other systems; positive values, shaded blue, indicate the system was preferred more often than not, and negative values, shaded red, the reverse. Grader options are frontier models serving as LLM-as-a-judge systems (GPT-5.5, Gemini 3.1 Pro, Claude Opus 4.8) and the human experts; gradees are GPT-5.5, Gemini 3.1 Pro, Claude Opus 4.8 and OpenEvidence (OE). The three cells highlighted along the diagonal of each panel indicate self-judging, where a model grades its own answers. As a specialist AI system, OE appears only as a system being rated, not as a grader. 95\% bootstrap CIs are shown.}
  \label{fig:grader_gradee_winrate_matrix_all_axes}
\end{figure}

One of the main contributions of this work is a large-scale gold standard specialist evaluation of LLM outputs. Such scaled human evaluation-based analyses are challenging due to the practical logistics of recruiting physician evaluators, and LLMs are instead commonly substituted as judges, with or without a rubric \citep{Arora2025-fp, Schmidgall2026-cg, Fast2024-gc}. The large-scale human evaluation in this study provides an opportunity to compare the two approaches. To assess how effectively zero-shot LLM-as-a-judges approximate ratings by clinical experts, each of the frontier models was prompted to complete the exact same head-to-head evaluation as the human specialists. The exact same experimental procedure was conducted, including the same questions, answers, and pairs that the humans saw. Answers were blinded and randomized in the same way as in the human expert evaluation for the text-only condition.

The results show that LLM judges generally agreed with human experts on which AI model performed the best, but significantly differed for the other ranks (Figure~\ref{fig:grader_gradee_winrate_matrix_all_axes}). Self-preference was notably high for GPT-5.5 \citep{Koo2024-gd, Zheng2023-wp}, particularly with respect to accuracy and clinical utility; consequently, its average correlation with the human expert ratings was low and, in fact, negative  (Kendall's tau=-0.200).
Ratings from Gemini 3.1 Pro and Claude Opus 4.8 were better aligned with those from clinical experts but still differed substantially in rankings, with average correlations of 0.467 and 0.267, respectively.
LLM-as-a-judge systems were also generally overconfident, reporting more extreme win differences compared to the clinical experts.
Finally, we note that even when using an ensemble of LLMs (also known as an LLM-as-a-jury \citep{Verga2024-vf, Chan2024-dj, Kalra_undated-rn}), average correlation remained low at -0.067.

\section{Discussion}\label{sec:discuss}

In this large, blinded, specialty-matched evaluation of clinical AI systems, practicing U.S. physicians rated answers to real-world queries sampled from a live point-of-care clinical decision support system (Real-POCQi).
A specialized clinical tool (OE) was rated as being more accurate, more clinically useful, better sourced, more easily verified, and more complete than three frontier general-purpose models (GPT-5.5, Gemini 3.1 Pro, and Claude Opus 4.8).
This result was robust in sensitivity analyses, where analyses were stratified across rendering modes with or without citations, longer versus shorter answers, evaluators who were or were not users of the specialized tool, and restriction to HealthBench questions \cite{Arora2025-fp}.
Given that the OE platform was specifically tuned to cater to its users, these results should not be overly surprising.
They do not mean that general-purpose models cannot answer such questions, but rather show how substantial engineering and customization can significantly improve performance and utility for a specific user base.

The evaluation of clinical AI tools requires the navigation of numerous tradeoffs: scalability versus involvement of human expertise, controlled simulation vs grounding in real world behaviors, tool blinding versus preservation of customized features, and enforcement of operative norms versus allowance for the flexibility in the hands of individual users.
In this study, we sought foremost to align both the topics and evaluation with the needs of domain experts who use a real-world AI-powered clinical decision support platform.
Thus, the benchmark evaluation used real-world questions and a multi-axis grading process performed by topic-matched specialists.
Strengths of this benchmark study included the recruitment of nearly 150 specialists across 30 medical specialties in the U.S. and five axes compared in a pairwise fashion, allowing physicians to use their own clinical judgment to evaluate AI model responses to real-world clinical queries.
Such an approach has been shown to be more accurate than rubric-based scoring systems, as the latter requires presupposed criteria \citep{Yang2026-pr}.
The consistency of results on HealthBench questions further supports generalizability beyond the Real-POCQi distribution.
Along similar lines, other recent works have also sought to incorporate more real-world data into clinical AI benchmarks \citep{Wu2026-bq}.

Our results contrast with recent findings from \citep{Vishwanath2026-io}, which reported generalist tools outperforming specialized systems including OE on several rated clinical dimensions. While a precise evaluation of points of disagreement in these studies is not is not possible because their query set is not yet publicly available, several differences in study design plausibly may contribute. Their question set was drawn from an enterprise ChatGPT instance whereas ours was drawn from questions on OpenEvidence. It is quite possible clinicians have different purposes for these two tools and interact with them in different ways, leading to discordance in measured performance.
Their study also involved many fewer physician graders (12 vs. 149 in this study) from a single institution, whereas this work involved graders spanning 36 states.
Their evaluation design also differed: their reviewers scored each answer in isolation on a four-point scale based on standardized rubrics with no known matching of specialist raters to question topic, whereas ours matched specialties to preserve clinical context as much as possible and asked physicians to compare paired answers.
Finally, the performance of general-purpose AI models is highly dependent on prompt engineering, and it is possible that individualized model-specific prompt-tuning could lead to frontier model improvements.
Disentangling the contributions of rater expertise, question selection, and evaluation design to this divergence is an important direction for future work.

Our comparisons between human experts and LLM-as-a-judge systems also highlight the need for caution when automating judgment of clinical answers.
When general-purpose AI models were prompted to grade answers, their rankings often differed from that of human experts and were also systematically overconfident. Self-preference was most pronounced when grading accuracy, presenting a direct threat to clinical validity. Alignment with domain experts varied by judge and evaluation dimension.
Agreement was highest when ranking completeness, as longer answers tended to be preferred along this axis by both humans and automated judges.
Judges also tended to agree on which AI system performed the best but disagreed on later ranks.
These divergences caution against assuming automated judges or juries are good approximations of real-world capabilities of AI systems; rather, substantial tuning and validation of LLM judges/juries must be done prior to using such mechanisms \citep{Chatzi2024-cy, Vossler2026-ii}.

In reconciling these results with prior works, one explanation may be that general-purpose models do well on general AI evaluation frameworks but not those designed to target specific deployment settings, such as our investigation of point-of-care clinical decision support.
At the same time, AI tools precisely tailored to target these settings can substantially outperform general-purpose AI models when evaluated in a fit-for-purpose benchmark, but these tools may not generalize to other specific deployment environments (e.g., \citep{Vishwanath2026-io}).
As such, a key question in designing AI evaluations is what a given benchmark dataset can and cannot measure or, more realistically, what it is better or worse at measuring.
This work too presents tradeoffs---the approach of matching physician graders to question topics optimizes evaluation along the accuracy axis, but it may lower the quality of evaluations on other dimensions such as clinical utility.
Thus, by releasing this benchmark dataset publicly, we hope it will help advance the science of clinical AI evaluation by providing a unique window into the real-world queries asked by physicians and move the field towards measuring the net impact of these tools on clinical judgment and, ultimately, patient outcomes.

\subsection*{Limitations}

Several limitations qualify these findings. The benchmark questions were sampled from a single week (June 7–14, 2026); although drawn from a pool of more than seven million queries submitted by physicians across the United States and the distribution of question themes generally remain stable over time, these queries capture one moment in a distribution and the exact questions asked will shift as practice patterns and tools evolve \citep{Finlayson2021-ad}.
As such Real-POCQi originate from the OE platform, its distribution may be shaped by routing — clinicians may direct certain kinds of questions to a specialized clinical tool — so the question mix could differ systematically from an independently constructed test set.
Another concern is that OE implemented data collection and conducted the survey, though the data collection approach was agreed upon with the authors who are unaffiliated with OE.
Nevertheless, sensitivity analyses suggest that the observed results are quite robust.
Findings on the HealthBench dataset were directionally consistent, OE's performance across different axes were uniformly positive, and the ordering among the general-purpose models remained generally stable.
We have provided the questions and answers for public review.

The ratings reflect the physicians who chose to respond to our survey, who may differ systematically from non-respondents and introduce participation bias.
Furthermore, while we matched questions to physicians by their clinical specialty, this may not sufficiently capture the clinical context in which the question was asked.
As such, a high-scoring question in our setting may not reflect the true score that would be given in the midst of care delivery.
Our evaluation also provides only a snapshot of current systems; model performance changes as systems are updated \citep{Feng2022-mk, van-Amsterdam2026-az}, and the cost of re-collecting expert ratings at that cadence is itself a barrier to keeping such benchmarks current.
We also presented the frontier models with each clinical query directly, with fairly limited clinical prompt engineering. Our findings should therefore be read as a comparison of the accuracy and form of these systems as a clinician would encounter them via a general user interface, rather than as a ceiling on the underlying models' capabilities.
Finally, while the number of questions and physician graders is large in aggregate, the benchmark dataset remains limited for estimating performance within any individual specialty and only a small number of models were evaluated.

\section{Conclusion}\label{sec:conclusion}

In a randomized, blinded, and specialty-matched evaluation, U.S. physicians rated answers to real clinical queries asked by practicing clinicians to a clinical AI tool, with consistent results on an independent benchmark.
Answers from a specialized clinical tool were consistently rated higher than those from three frontier general-purpose models across every dimension assessed.
These results highlight that understanding the true capabilities of AI tools in their deployment environment requires evaluations that are fit-for-purpose, which requires sampling real-world queries ``in the wild'' \citep{Lin2025-kr} and designing evaluations to be as reflective of the query's original context as much as possible.
We share the full set of questions, answers, and ratings in this benchmark dataset to the research community, with the hopes of advancing the science of clinical AI evaluation.

\backmatter

\section{Methods}
\label{sec:methods}

\subsection{Study Overview}

\begin{figure}
    \centering
    \includegraphics[width=0.35\linewidth]{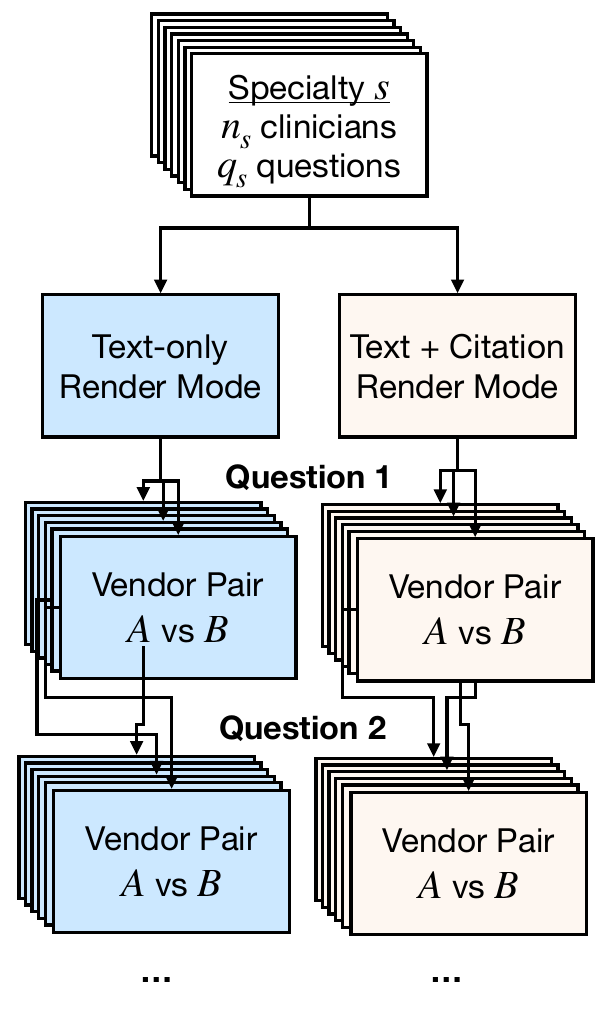}
    \caption{
\textbf{Schematic of the nested randomization design.}
    Clinical questions and physician evaluators are sampled and matched by specialty $s$ ($q_s$ questions and $n_s$ clinicians per specialty).
    Each clinician who begins the survey is randomized to a single rendering mode that is held fixed throughout --- text-only (blue) or text-with-citations (tan).
    For every assigned question, a pair of the four AI systems (vendor pair $A$ vs $B$) are randomly drawn and the two answers are presented in a randomized A/B order.
}
    \label{fig:randomized_study}
\end{figure}

We conducted a randomized mixed factorial design with blinded pairwise model evaluations to benchmark the win difference between answers from four LLM candidates---GPT-5.5, Gemini 3.1 Pro, Claude Opus 4.8, and OpenEvidence (OE)---on real-world clinical questions submitted to the OE platform.
Physicians rated responses from randomly selected LLM pairs on questions whose clinical topic matched their specialty (Figure~\ref{fig:randomized_study}).

\subsection{Study Population}

The study population was characterized by both the population of clinical decision support queries originating from OE users as well as the population of evaluators.

Queries were sampled from real user queries submitted to the OE platform between June 7--14, 2026 as well as HealthBench \citep{Arora2025-fp}.
The following exclusion criteria were applied: (1) questions potentially containing protected health information (PHI); (2) questions asked mid-conversation (i.e., follow-up turns); (3) non-English queries; and (4) non-clinical-decision-support tasks such as translation, documentation, or summarization.
For the OE queries that passed the filter, 100 questions were sampled across each of the 36 medical specialties from US physicians.
Sampling was stratified 50/50 between questions without patient context and questions containing patient information.
All OE questions were lightly rewritten using Opus-4.6 to remove any identifiable information and abide by user agreements while retaining the clinical question content (prompt in Figure~\ref{fig:question_rewrite_system_prompt}).

Evaluators were required to be licensed, actively practicing physicians in the U.S.. Physicians were recruited from IQVIA's OneKey provider database, a healthcare data and analytics company whose provider reference database includes contact and self-reported specialty information for the large majority of licensed U.S. physicians holding a valid National Provider Identifier (NPI). Each physician's specialty was identified from the National Plan and Provider Enumeration System, the national NPI registry, and evaluators were matched to questions corresponding to their specialty.

To ensure the evaluator pool reflected the clinical content under assessment, the recruitment sample was constructed to mirror the specialty distribution of the question set: each specialty's share of the physicians contacted was set equal to its share of the questions, yielding a final sample that matched this target distribution to within approximately 0.2 percentage points. Recruitment was stratified by prior platform exposure, distinguishing physicians active on OpenEvidence (``On-OE") from those who had never held an OpenEvidence account (``Off-OE").

Between June 16 and June 21, 2026, OE emailed a total of 12,000 physicians (5,025 On-OE and 6,975 Off-OE).
Of these, 214 responded (121 On-OE, 93 Off-OE), an overall response rate of 1.8\% (2.4\% among On-OE physicians and 1.3\% among Off-OE physicians).
A total of 149 physicians completed the full blinded evaluation (78 On-OE, 71 Off-OE), corresponding to a completion rate of 1.2\% relative to those contacted (1.6\% On-OE, 1.0\% Off-OE) and 69.6\% relative to those who responded. Eligible clinicians who completed the evaluation were compensated for their participation.

\subsection{Evaluation Dimensions and Scale}

Evaluators assessed their preference for each answer pair across five dimensions:
\begin{enumerate}
    \item \textbf{Accuracy}: Which response is more factually accurate?
    \item \textbf{Clinical utility}: Which response is more useful for providing high-quality clinical care?
    \item \textbf{Source quality}: Which response is supported by more authoritative source material?
    \item \textbf{Verifiability}: Which answer is easier to verify?
    \item \textbf{Completeness}: Which response more completely addresses the question?
\end{enumerate}

For each dimension, evaluators rated their preference on a 5-point Likert scale: A is much better, A is somewhat better, About equal, B is somewhat better, and B is much better.
Each dimension was rated independently.
Options were mutually exclusive with no neutral default pre-selected.

\subsection{Randomization and Blinding}

Randomization occurred at multiple levels:
\begin{enumerate}
    \item \textbf{Rendering mode assignment.} Upon signup, each evaluator was randomly assigned to one of two rendering modes---text-only (primary analysis) or text-with-citations (secondary analysis)---held fixed for all their assigned questions.
    The text-with-citations mode was designated as secondary because citations may make the vendor identity less blinded, as the cited sources and citation style of an answer can differ systematically across AI systems, potentially allowing raters to infer which system produced it or influencing their quality judgments based on those sources.
    \item \textbf{Question assignment.} The system selected specialty-matched questions uniformly at random from the question bank and assigned each evaluator up to 18 questions.
    \item \textbf{LLM pair and position assignment.} For each assigned question, a sequence of two LLM candidates was randomly sampled without replacement from the four systems, such that the assignment to position A versus position B was randomized. OE was assigned a sample weight of 1/3 to power comparisons to other models. Unequal sample weights were adjusted for in the statistical analysis.
\end{enumerate}

\begin{figure}[tbp]
  \centering
  \includegraphics[width=0.9\linewidth]{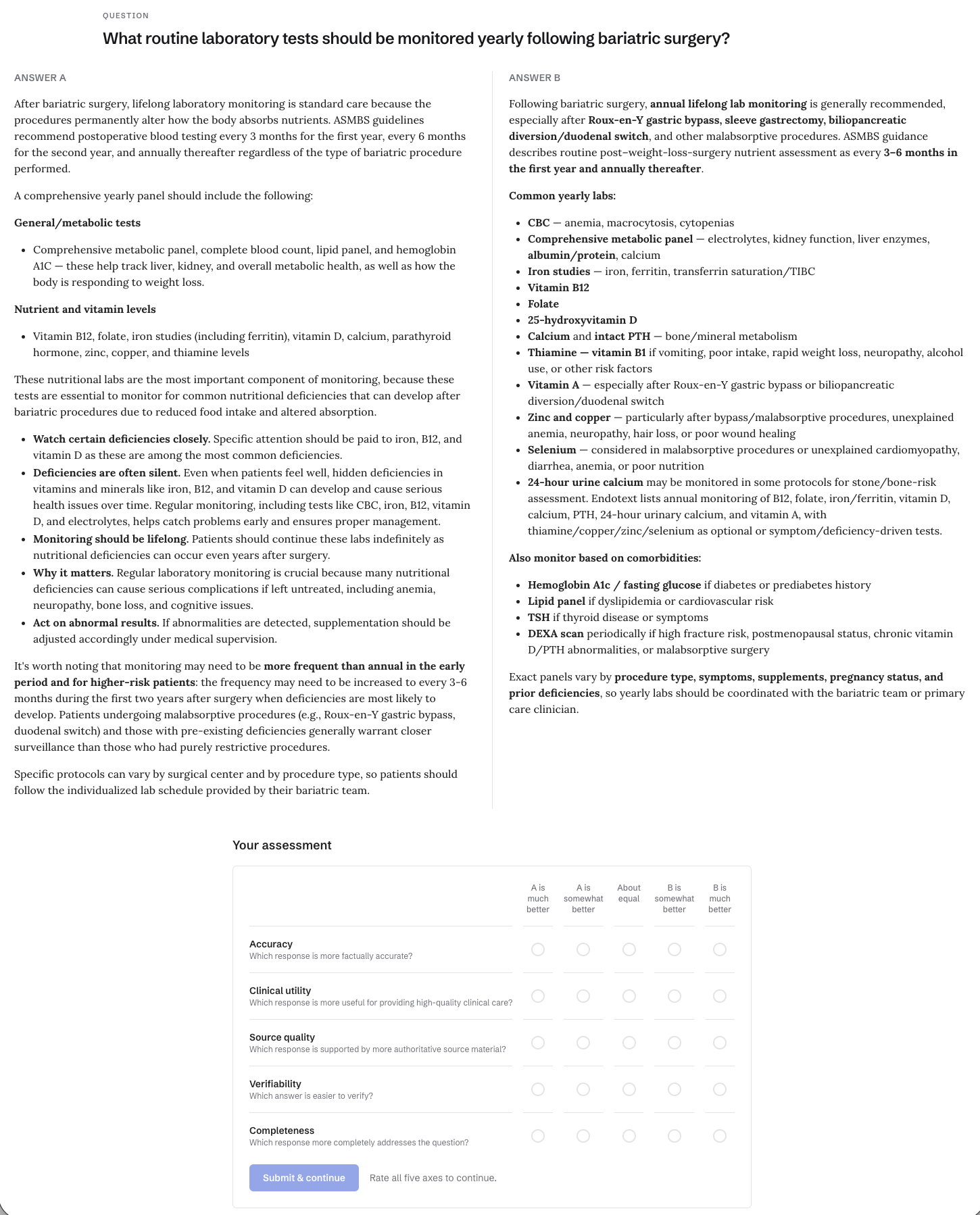}
  \caption{\textbf{Blinded pairwise rating interface.}
  The clinical question appears at the top, with two AI system answers displayed side by side as \emph{Answer A} and \emph{Answer B}.
  For each comparison, a random pair of the four systems was drawn and assigned to A/B positions at random.
  Vendor identities were hidden and both answers were rendered in identical formatting so that systems could not be distinguished by visual style.
  Evaluators rate the pair on five dimensions using a five-point preference scale with no option pre-selected.}
  \label{fig:rating_interface}
\end{figure}

The identity of the LLM vendor was blinded throughout the study.
Evaluators were not informed which systems were being compared and the order in which randomly selected LLM candidates were presented to raters was also randomly assigned.
A custom display system rendered both answers identically within a given evaluator's rendering mode, preventing identification by visual style (Figure~\ref{fig:rating_interface}).
Statistical analyses were fully pre-specified, with LLM identities blinded throughout the analysis.

\subsection{LLM Queries}

All systems were queried programmatically via their respective APIs.
The OpenEvidence API was queried for the version available on June 15 2026 (no version number is available).
For OpenEvidence in particular, API access was used to avoid formatting and truncation artifacts that can arise from copying responses out of the web interface.
To maximize reproducibility, frontier-model generations used deterministic settings throughout: temperature was fixed at 0.0 to remove sampling variability, and a fixed random seed of 42 was applied across all runs.
Thinking was automatically determined by the LLM.
Web search was enabled for all frontier models.

\subsection{Statistical Analysis}

All statistical analyses were pre-specified prior to data analysis and all data were accessible only by an independent statistician (JF).
Prior to analysis, LLM candidate identities were randomized and blinded.

\subsubsection{Definitions}

For a given evaluation dimension $D$, we define the \textit{win-rate} as $D_{A,B} = P(A \succ B)$ between LLM $A$ and LLM $B$, where $A \succ B$ indicates that $A$ is rated ``somewhat better'' or ``much better'' than $B$.
The \textit{win-rate matrix} contains these values for all possible pairs.
We define the \textit{win difference} (also known as net benefit, win-rate difference, or favorability score~\citep{Pocock2012-md,Fang2026-cv, Lavrakas2008-vz}) as:
\begin{equation}
\Delta D_{A,B} = P(A \succ B) - P(B \succ A)
\end{equation}
representing the percentage of wins minus percentage of losses.
For one-vs-rest (OvR) win difference, $\Delta D_{A,\text{rest}}$ is defined as the difference between the uniform mixture of win rates, i.e., where the comparator system $B$ is sampled uniformly at random among the remaining LLMs.
To adjust for uneven sampling of LLM vendors, the win difference was estimated by reweighting observations such that all pairs have equal weights.
Note that head-to-head win rates do not require such adjustments.

\textit{Inter-rater agreement} was evaluated between pairs of raters presented with the same question, LLM pair, and render mode, where agreement was defined as rating within one Likert unit.

All confidence intervals were reported at the 95\% level and computed using the bootstrap with question-level clusters \citep{Miller2024-ox, Vossler2026-ii}.
All reported p-values were two-sided and the threshold for statistical significance was set at $p = 0.05$.
(We note that despite this study conducting a large number of hypotheses tests, the central p-values are only those from the primary analysis.
The remaining p-values are those from sensitivity analyses, which are simply to test the robustness of the findings.)

\textbf{Inclusion/Exclusion.}
Question ratings completed in under 10 seconds were excluded from analysis to ensure evaluators had sufficient time to review both responses.
Partial sessions contributed their completed comparisons.
Evaluators who completed some but not all of their assigned questions had their completed comparisons included in the analysis.

\subsubsection{Primary Analysis}

The primary analysis compared the OvR win difference for each LLM across all five evaluation dimensions within the text-only rendering mode.
For each evaluation dimension $D$ and LLM $A$, we tested the null hypothesis that the win difference is zero, i.e., $H_0: \Delta D_{A,\text{rest}} = 0$.
The p-value was obtained using a permutation test, where the winner of each LLM pair was randomly flipped with 50\% (ties remain unchanged).
The confidence interval was obtained by resampling questions.

\subsubsection{Secondary Analyses}

We conduct two secondary analyses.
First, to increase the granularity of the LLM comparison, we estimated the full win-rate matrix between all possible pairs of systems for each evaluation dimension, pooling across the text-only and text+citation rendering conditions.
Second, we repeated the primary analysis but for the text-with-citations rendering mode.

\subsubsection{Sensitivity Analysis}

We conducted sensitivity analyses to evaluate the robustness of our findings from the primary and secondary analyses.

To check robustness to how wins were defined, the primary analysis is rerun but with a stricter win/loss definition, where winning required ``much better'' and losing required ``much worse,'' with ``somewhat better/worse'' counted as ties.

The length of a model’s response to a clinical query may alone influence clinician ratings of that response (length bias). To evaluate the potential impact of length bias in the evaluation of LLM systems \citep{Zheng2023-wp, Koo2024-gd}, we stratified results from the primary analysis according to whether the evaluated system's response was longer shorter than its comparator.

To assess generalizability beyond the Real-POCQi distribution, we repeated the primary analysis restricted to questions drawn from HealthBench \citep{Arora2025-fp}.

Finally, we evaluated the robustness of our findings to correlation induced by physicians grading multiple questions.
By instead viewing the benchmark questions as fixed but the graders as samples from a larger population, we recomputed $p$-values and confidence intervals using the cluster bootstrap but with clusters defined by the grader rather than the question.

\subsubsection{Exploratory Analysis}

Rankings between the four LLMs were estimated using a Davidson model that allowed for ties in pairwise comparisons \citep{Davidson1970-hf} (Appendix Figure~\ref{fig:davidson_worth_by_model}).
(We refer to this analysis as exploratory given the parametric assumptions typically used in ranking models.)
Alignment between the human expert and LLM-as-a-judge/jury were quantified in terms of Kendall's tau, since the number of rankings was small.

Themes of the clinical queries in the benchmark suite were identified using an LLM-based unsupervised clustering method.
Following a workflow based on HACHI~\citep{Feng2026-ka}, an LLM (Claude Sonnet 4) first brainstormed keyphrases from each question, the keyphrases were clustered using Bertopic \citep{Grootendorst2022-it} and then interpreted as candidate themes by an LLM (Gemini), human experts then reviewed and edited the themes, and an LLM (Claude Sonnet 4) finally tagged every question based on the discovered themes.

\subsection{LLM-as-Judge Comparison}

To complement the expert human analysis, we conducted a parallel LLM-as-judge evaluation using the three frontier LLMs as a judge, each replicating exactly the expert human process. For each answer pair that a human evaluated, each LLM judge was given the same answer pair where the positions were randomized and the LLM identities were blinded.
The LLM judges were asked to rate A vs B on the same five axes used by the human evaluators.
The prompt used for each LLM judge is shown in Figure~\ref{fig:machine_grader_system_prompt}.
Answers were serialized into markdown format using the same serialization code, and the question and answers were passed to each LLM in a user message.
The thinking mode was ``minimal'' and temperature was 0. The result was four fully parallel datasets of ratings for the same questions and answers, one graded by each of the three LLM judges and one graded by the expert human specialists.

\subsection{Ethics}

This study received an exempt determination from Rapid IRB (IRB number IRB00014585; Federalwide Assurance FWA00035509) under the study title ``Expert user evaluation of chat-based medical AI tools.'' All principal investigators and study personnel completed IRB human-subjects training.

\subsection{Conflicts of Interest}

OpenEvidence (OE) is one of the systems compared in the evaluation.
OE (points of contact: Samuel G. Finlayson, Travis I. Zack, and Zachary Ziegler) co-developed the data collection plan with JF and YM, implemented data collection, and paid survey responders.
None of the authors of this study have an affiliation with OpenEvidence.
JF, YM, and PH fully pre-specified the statistical analysis plan and JF conducted all statistical analyses with full blinding to LLM identity.

\subsection{Data and Code Availability}

All questions, answers, and ratings have been publicly released at \url{https://huggingface.co/datasets/jjfenglab/Real-POCQi}.
All code for the pre-specified statistical analysis is available at \url{https://github.com/jjfenglab/Real-POCQi-statistics}.

\begin{appendices}
\renewcommand{\theHfigure}{\Alph{section}\arabic{figure}} 
\renewcommand{\theHtable}{\Alph{section}\arabic{table}}

\section{More dataset details}

\begin{figure}
  \centering
  \includegraphics[width=0.7\linewidth]{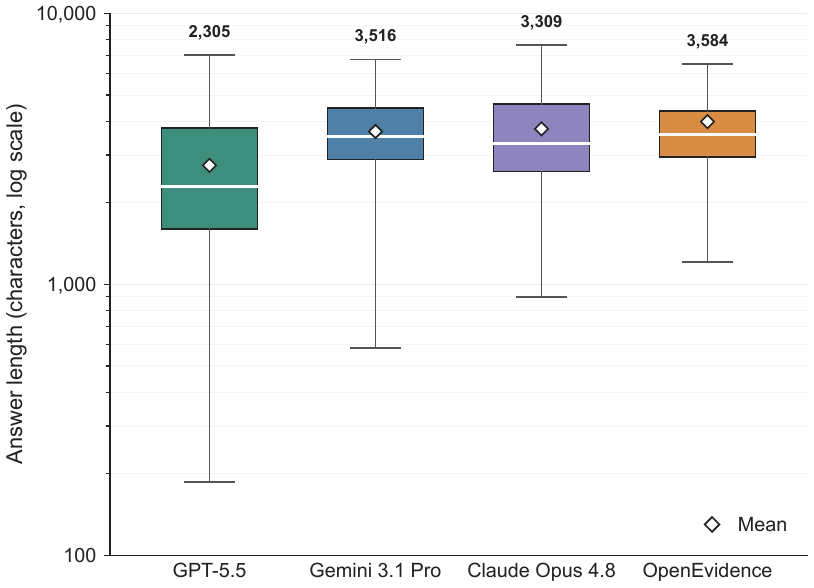}
  \caption{\textbf{Answer length distribution by AI system.} Distribution of answer length, in characters, for each of GPT-5.5, Gemini 3.1 Pro, Claude Opus 4.8 and OpenEvidence. Boxes span the interquartile range (IQR) with the median as the white line, whiskers extend to the most extreme answer within 1.5$\times$IQR of the quartiles. The open diamond marks the mean and the value above each box is the median.}
  \label{fig:answer_length_distribution}
\end{figure}

Additional detailed statistics describing the dataset used for the evaluation are shown in Table~\ref{tab:descriptives}. The distribution of answer length for each system analyzed is shown in Figure~\ref{fig:answer_length_distribution}.

\begin{table}
  \centering
  \footnotesize
  \setlength{\tabcolsep}{6pt}
  \renewcommand{\arraystretch}{1.2}
  \caption{\textbf{Descriptive statistics for the benchmark dataset and evaluator cohort.}
}
  \label{tab:descriptives}
  \begin{tabularx}{\linewidth}{@{}>{\raggedright\arraybackslash}p{5.5cm} >{\centering\arraybackslash}X >{\centering\arraybackslash}X@{}}
    \toprule
    Metric & Real-POCQi & HealthBench \\
    \midrule
    Total questions & 620 & 187 \\
    Total ratings collected & 1156 & 389 \\
    Specialties & 30 & 28 \\
    Graders & 149 & 130 \\
    Number of U.S. states covered by graders & 36 & 34 \\
    Num ratings per question (mean $\pm$ SD) & 1.9 $\pm$ 1.3 & 2.1 $\pm$ 1.5 \\
    Ratings collected per LLM & OE: 742; GPT-5.5: 508; Claude Opus 4.8: 539; Gemini 3.1 Pro: 523 & OE: 260; GPT-5.5: 175; Claude Opus 4.8: 180; Gemini 3.1 Pro: 163 \\
    Graders with $\geq$1 rating per rendering mode & text-only: 100; text+citations: 49 & text-only: 86; text+citations: 44 \\
    Completed assignments per grader (mean $\pm$ SD) & 7.8 $\pm$ 4.0 & 3.0 $\pm$ 1.8 \\
    Survey completion rate per grader (mean $\pm$ SD) & 58.1\% $\pm$ 26.6\% & 22.3\% $\pm$ 13.9\% \\
    Submission time, seconds (mean $\pm$ SD) & 291.5 $\pm$ 1641.8 & 231.2 $\pm$ 320.5 \\
    \bottomrule
  \end{tabularx}
\end{table}

\section{Sample questions}

A random sample of 20 questions, along with the specialty associated with the question is shown in Table~\ref{tab:sample_questions}.

\begingroup
\footnotesize
\setlength{\LTleft}{0pt}\setlength{\LTright}{0pt}
\setlength{\LTcapwidth}{\linewidth}
\setlength{\extrarowheight}{2pt}
\begin{longtable}{@{}>{\raggedright\arraybackslash}p{0.5cm}>{\raggedright\arraybackslash}p{2.6cm}>{\raggedright\arraybackslash}p{\dimexpr\linewidth-3.1cm-4\tabcolsep\relax}@{}}
\caption{\textbf{Sample questions from the OpenEvidence evaluation dataset.} A random sample of 20 of the OpenEvidence-derived questions used in the evaluation, shown with the clinical specialty assigned to each question and sorted by specialty.}\label{tab:sample_questions}\\
\toprule
\textbf{\#} & \textbf{Specialty} & \textbf{Question}\\
\midrule
\endfirsthead
\multicolumn{3}{@{}l}{\footnotesize\itshape Table~\ref{tab:sample_questions} continued}\\
\toprule
\textbf{\#} & \textbf{Specialty} & \textbf{Question}\\
\midrule
\endhead
\midrule
\multicolumn{3}{r@{}}{\footnotesize\itshape continued on next page}\\
\endfoot
\bottomrule
\endlastfoot
1 & Dermatology & What is the management of hidradenitis suppurativa abscess?\\
\addlinespace[2pt]
2 & Dermatology & What topical antifungal agent is most appropriate for treating candidal intertrigo in a skin fold that contains a chronic non-infected draining wound being managed by wound care?\\
\addlinespace[2pt]
3 & Endocrinology & What happens to free T3 levels during pregnancy?\\
\addlinespace[2pt]
4 & Gastroenterology & Should routine H. pylori testing be performed in all patients diagnosed with gastric cancer?\\
\addlinespace[2pt]
5 & Geriatrics & Does statin use influence cognitive decline patterns in older adults?\\
\addlinespace[2pt]
6 & Infectious Disease & Do you need to identify the subtype of MAC to initiate treatment?\\
\addlinespace[2pt]
7 & Infectious Disease & Under what circumstances should testing for chromosomally integrated HHV-6 be performed?\\
\addlinespace[2pt]
8 & Neurology & What causes nausea as a side effect of dopamine agonists used in the treatment of Parkinson's disease?\\
\addlinespace[2pt]
9 & OB / GYN & Management of pelvic inflammatory disease\\
\addlinespace[2pt]
10 & OB / GYN & How soon after delivery should HbA1c be rechecked in patients with prediabetes?\\
\addlinespace[2pt]
11 & OB / GYN & Is oral fluconazole contraindicated in pregnancy, and if yes, what is the preferred alternative treatment?\\
\addlinespace[2pt]
12 & OB / GYN & What is the appropriate management for a biophysical profile score of 4/10 with adequate amniotic fluid volume?\\
\addlinespace[2pt]
13 & Oncology / Hematology & What percentage of patients with N2 rectal cancer were enrolled in the OPRA trial?\\
\addlinespace[2pt]
14 & Oncology / Hematology & What liver enzyme elevation is commonly associated with nab-paclitaxel administration?\\
\addlinespace[2pt]
15 & Oncology / Hematology & What types of infections are particularly dangerous or commonly seen in patients with hereditary spherocytosis?\\
\addlinespace[2pt]
16 & Ophthalmology & Is central serous chorioretinopathy associated with semaglutide use?\\
\addlinespace[2pt]
17 & Psychiatry & How does postpartum depression in mothers influence the likelihood of depression developing in their children?\\
\addlinespace[2pt]
18 & Pulmonology & What size distinguishes a pulmonary nodule from a pulmonary mass?\\
\addlinespace[2pt]
19 & Pulmonology & Is administering a pneumococcal vaccine to patients with underlying respiratory conditions like COPD or asthma when more than 5 years have passed since their last dose consistent with current recommendations?\\
\addlinespace[2pt]
20 & Surgery & What are all the anastomoses at the foot involving the tibial vessels?\\
\addlinespace[2pt]
\end{longtable}
\endgroup

\section{LLM Prompts}

The prompt used to rewrite user questions on OpenEvidence to ensure exact verbatim questions were avoided is shown in Figure~\ref{fig:question_rewrite_system_prompt}. Claude Opus 4.6 was used to do the rewriting with temperature 0.

The prompt used for the LLM-as-judge data is shown in Figure~\ref{fig:machine_grader_system_prompt}. This prompt was used for all frontier model judges, with temperature 0.

\begingroup
\lstset{
  basicstyle=\ttfamily\footnotesize,
  breaklines=true,
  breakatwhitespace=false,
  postbreak=\mbox{\textcolor{gray}{$\hookrightarrow$}\space},
  columns=fullflexible,
  keepspaces=true,
  frame=single,
  framesep=6pt,
  rulecolor=\color{black!40},
  backgroundcolor=\color{black!2},
  xleftmargin=4pt,
  xrightmargin=4pt,
  aboveskip=0pt,
  belowskip=0pt,
}
\begin{figure}[tbp]
  \centering
  \begin{lstlisting}
You are a medical question rewriter. Given a medical question, rewrite it as a slightly different, but equivalent question. Rules:

- The core meaning, difficulty, and clinical reasoning required must remain the same
- Keep the same format as the original (e.g., if it's a short question, output a short question)
- Do NOT turn the question into a multiple-choice question or add answer options
- Do NOT add clinical vignettes, patient scenarios, or additional patient details if the original does not have them
- Change only VERY minor details
- The rewritten question must not be a verbatim copy of the original

Only output the rewritten question. Say nothing else.
  \end{lstlisting}
  \caption{\textbf{Question-rewriting system prompt used to build the evaluation dataset.} The verbatim system prompt given to the model that rewrites each OpenEvidence query into a slightly altered but clinically equivalent variant before it enters the dataset.}
  \label{fig:question_rewrite_system_prompt}
\end{figure}
\endgroup

\begingroup
\lstset{
  basicstyle=\ttfamily\footnotesize,
  breaklines=true,
  breakatwhitespace=false,
  postbreak=\mbox{\textcolor{gray}{$\hookrightarrow$}\space},
  columns=fullflexible,
  keepspaces=true,
  frame=single,
  framesep=6pt,
  rulecolor=\color{black!40},
  backgroundcolor=\color{black!2},
  xleftmargin=4pt,
  xrightmargin=4pt,
  aboveskip=0pt,
  belowskip=0pt,
}
\begin{figure}[tbp]
  \centering
  \begin{lstlisting}
You are an expert US physician serving as an evaluator of a clinical decision support system that is used by physicians in a real clinical setting.
You will be given a question and two answers, labelled "Answer A" and "Answer B". Judge the answers specifically in the context of the user being a physician, not a patient.

Rate which answer is better on EACH of these five axes independently:
- clinical_utility: Which response is more useful for providing high-quality clinical care?
- accuracy: Which response is more factually accurate?
- source_quality: Which response is supported by more authoritative source material?
- verifiability: Which answer is easier to verify?
- completeness: Which response more completely addresses the question?

For each axis choose exactly one option from this 5-point scale:
- "strongly_a": Answer A is much better
- "slightly_a": Answer A is somewhat better
- "tie": about equal
- "slightly_b": Answer B is somewhat better
- "strongly_b": Answer B is much better

Use the full scale; reserve "tie" for genuinely indistinguishable answers.
Reply with ONLY a JSON object whose keys are the five axis names above and
whose values are the chosen scale option strings, e.g.:
{"clinical_utility":"slightly_a","accuracy":"tie","source_quality":"strongly_b","verifiability":"slightly_b","completeness":"slightly_a"}
Always output the complete JSON object, do not truncate it.
  \end{lstlisting}
  \caption{\textbf{LLM-as-judge prompt used in the machine-grading experiment.} The verbatim system prompt given to each of the three frontier models (GPT-5.5, Gemini 3.1 Pro, Claude Opus 4.8) acting as automated raters. The per-pair question and the two answer texts were supplied in the user message.}
  \label{fig:machine_grader_system_prompt}
\end{figure}
\endgroup

\section{Additional Results}

\begin{table}[tbp]
  \footnotesize
  \setlength{\tabcolsep}{4pt}
  \renewcommand{\arraystretch}{1.15}
  \caption{\textbf{One-vs-rest win differences across the primary, secondary, and sensitivity analyses.}
  One-vs-rest win difference (\%) for each AI system (OpenEvidence, GPT-5.5, Claude Opus 4.8, Gemini 3.1 Pro) on each of the five evaluation axes.
  The primary analysis used text-only rendering and the secondary analysis used text-with-citations.
  Sensitivity analyses re-estimate win differences under (i) restriction to HealthBench questions, (ii) a strict win/loss definition counting only ``much better/worse'' ratings as wins/losses, (iii) stratification by whether the evaluated system's answer was longer or shorter than its comparator, and (iv) stratification by whether the evaluating physician was a registered OpenEvidence user.
  Values are reported in the format estimate [95\% bootstrap CI]; *$p<0.05$, **$p<0.01$, ***$p<0.001$ from the permutation test.
  }
\begin{tabular}{@{}l l r@{\,}l r@{\,}l r@{\,}l r@{\,}l@{}}
    \toprule
    \textbf{Analysis} & \textbf{Axis} & \multicolumn{2}{c}{\textbf{OE}} & \multicolumn{2}{c}{\textbf{GPT-5.5}} & \multicolumn{2}{c}{\textbf{Claude}} & \multicolumn{2}{c}{\textbf{Gemini}} \\
    \midrule
    Primary: & Accuracy & 24.7 & [18.4, 30.8]*** & $-$21.1 & [$-$29.1, $-$13.3]*** & $-$1.5 & [$-$9.3, 6.3] & $-$2.1 & [$-$9.5, 5.3] \\
    text-only & Clinical util. & 29.6 & [21.5, 37.4]*** & $-$19.4 & [$-$30.8, $-$8.5]*** & $-$13.7 & [$-$23.9, $-$3.5]** & 3.5 & [$-$7.0, 14.1] \\
    & Source Qual. & 38.8 & [31.7, 45.8]*** & $-$22.1 & [$-$32.4, $-$12.1]*** & $-$14.1 & [$-$23.7, $-$4.6]** & $-$2.7 & [$-$12.1, 6.4] \\
    & Completeness & 30.9 & [23.0, 38.6]*** & $-$30.2 & [$-$39.6, $-$20.7]*** & $-$3.8 & [$-$13.3, 5.7] & 3.1 & [$-$7.0, 12.9] \\
    & Verifiability & 26.2 & [19.5, 32.8]*** & $-$18.0 & [$-$26.4, $-$9.7]*** & $-$8.6 & [$-$16.6, $-$0.6]* & 0.5 & [$-$7.3, 8.1] \\
    \addlinespace[3pt]
    \midrule
    Secondary: & Accuracy & 35.7 & [27.8, 43.5]*** & $-$15.3 & [$-$26.5, $-$4.1]** & $-$11.2 & [$-$22.7, 0.5] & $-$9.3 & [$-$21.0, 2.4] \\
    text + cit. & Clinical util. & 46.2 & [36.2, 55.8]*** & $-$11.5 & [$-$27.4, 4.4] & $-$23.8 & [$-$37.8, $-$9.2]** & $-$10.9 & [$-$26.8, 5.0] \\
    & Source Qual. & 59.3 & [51.0, 67.3]*** & $-$21.4 & [$-$36.3, $-$6.7]** & $-$30.1 & [$-$42.4, $-$17.3]*** & $-$7.8 & [$-$22.5, 6.8] \\
    & Completeness & 51.6 & [41.9, 60.8]*** & $-$27.7 & [$-$42.7, $-$12.2]*** & $-$15.6 & [$-$29.3, $-$2.0]* & $-$8.2 & [$-$24.5, 7.5] \\
    & Verifiability & 59.7 & [51.2, 67.8]*** & $-$5.7 & [$-$20.7, 8.6] & $-$34.0 & [$-$46.8, $-$20.9]*** & $-$20.0 & [$-$34.3, $-$5.7]** \\
    \addlinespace[3pt]
    \midrule
    Sensitivity: & Accuracy & 14.1 & [3.3, 25.3]* & $-$8.0 & [$-$21.4, 5.5] & $-$9.1 & [$-$20.4, 2.4] & $-$2.9 & [$-$18.5, 11.6] \\
    HealthBench & Clinical util. & 24.7 & [10.0, 39.1]*** & $-$8.9 & [$-$26.9, 9.6] & $-$22.7 & [$-$38.6, $-$5.9]** & $-$2.0 & [$-$19.8, 15.9] \\
    & Source Qual. & 27.6 & [16.1, 39.1]*** & $-$18.8 & [$-$33.6, $-$3.5]* & $-$15.9 & [$-$30.1, $-$0.8]* & $-$4.9 & [$-$20.0, 10.3] \\
    & Completeness & 18.8 & [5.7, 32.1]** & $-$18.8 & [$-$35.4, $-$2.6]* & $-$11.4 & [$-$26.9, 4.5] & 3.9 & [$-$13.8, 21.0] \\
    & Verifiability & 25.3 & [13.8, 36.7]*** & $-$17.0 & [$-$30.6, $-$2.7]* & $-$12.9 & [$-$25.2, 0.0]* & $-$6.9 & [$-$20.8, 6.7] \\
    \addlinespace[3pt]
    \midrule
    Sensitivity: & Accuracy & 9.6 & [5.4, 13.9]*** & $-$12.6 & [$-$18.1, $-$7.3]*** & 2.8 & [$-$2.1, 7.8] & 0.2 & [$-$4.9, 5.3] \\
    strict & Clinical util. & 19.9 & [14.0, 25.6]*** & $-$18.3 & [$-$26.6, $-$10.1]*** & $-$3.5 & [$-$10.5, 3.7] & 1.8 & [$-$5.1, 8.7] \\
     & Source Qual. & 19.0 & [13.8, 24.0]*** & $-$16.7 & [$-$23.3, $-$10.2]*** & $-$1.2 & [$-$6.9, 4.5] & $-$1.1 & [$-$7.1, 4.7] \\
    & Completeness & 17.3 & [12.0, 22.4]*** & $-$21.9 & [$-$29.0, $-$14.7]*** & 4.6 & [$-$1.8, 11.0] & 0.0 & [$-$6.4, 6.4] \\
    & Verifiability & 11.8 & [7.5, 16.1]*** & $-$11.0 & [$-$16.8, $-$5.5]*** & 3.3 & [$-$1.9, 8.7] & $-$4.1 & [$-$8.9, 0.6] \\
    \addlinespace[3pt]
    \midrule
    Length: & Accuracy & 35.9 & [29.5, 42.1]*** & 2.1 & [$-$13.7, 19.1] & 7.1 & [$-$1.4, 16.0] & 6.1 & [$-$1.4, 13.6] \\
    longer & Clinical util. & 43.6 & [35.4, 51.6]*** & 10.4 & [$-$12.2, 32.9] & $-$3.7 & [$-$15.2, 7.9] & 9.1 & [$-$1.6, 19.8] \\
    & Source Qual. & 55.6 & [48.6, 62.0]*** & 4.6 & [$-$15.0, 25.4] & $-$0.9 & [$-$11.7, 9.8] & 6.0 & [$-$4.0, 16.0] \\
    & Completeness & 54.6 & [47.3, 61.7]*** & 29.2 & [11.4, 46.7]** & 16.1 & [5.5, 26.7]** & 18.4 & [8.5, 28.6]*** \\
    & Verifiability & 43.5 & [36.8, 50.0]*** & 8.0 & [$-$7.2, 23.4] & $-$6.3 & [$-$16.1, 3.4] & $-$0.0 & [$-$8.6, 8.5] \\
    \addlinespace[3pt]
    \midrule
    Length: & Accuracy & 17.2 & [9.3, 25.0]*** & $-$23.2 & [$-$30.1, $-$16.4]*** & $-$20.9 & [$-$30.8, $-$10.9]*** & $-$25.5 & [$-$36.0, $-$15.2]*** \\
    shorter & Clinical util. & 22.3 & [11.3, 33.5]*** & $-$22.7 & [$-$32.8, $-$13.0]*** & $-$34.8 & [$-$46.9, $-$22.3]*** & $-$20.3 & [$-$34.4, $-$6.2]** \\
    & Source Qual. & 30.6 & [20.9, 40.1]*** & $-$26.8 & [$-$36.0, $-$17.5]*** & $-$44.3 & [$-$54.1, $-$34.5]*** & $-$25.3 & [$-$37.9, $-$12.5]*** \\
    & Completeness & 12.1 & [1.4, 22.5]* & $-$41.4 & [$-$49.6, $-$33.0]*** & $-$39.1 & [$-$50.2, $-$27.7]*** & $-$37.1 & [$-$49.8, $-$23.9]*** \\
    & Verifiability & 28.1 & [19.2, 36.7]*** & $-$18.0 & [$-$25.9, $-$9.9]*** & $-$32.0 & [$-$41.7, $-$22.1]*** & $-$18.7 & [$-$29.7, $-$8.0]** \\
    \addlinespace[3pt]
    \midrule
    OE users & Accuracy & 33.8 & [27.3, 40.4]*** & $-$21.4 & [$-$29.9, $-$13.2]*** & $-$5.8 & [$-$14.7, 3.2] & $-$4.4 & [$-$13.1, 3.9] \\
    only & Clinical util. & 41.2 & [32.9, 49.8]*** & $-$16.2 & [$-$28.7, $-$4.4]** & $-$18.4 & [$-$29.5, $-$7.0]** & $-$4.3 & [$-$16.0, 6.9] \\
    & Source Qual. & 46.7 & [38.9, 54.1]*** & $-$19.9 & [$-$31.6, $-$8.4]*** & $-$16.8 & [$-$27.4, $-$6.3]** & $-$7.2 & [$-$17.5, 3.0] \\
    & Completeness & 44.3 & [35.8, 52.3]*** & $-$32.8 & [$-$42.9, $-$22.4]*** & $-$13.7 & [$-$24.1, $-$3.4]* & 4.4 & [$-$7.1, 15.4] \\
    & Verifiability & 40.6 & [33.0, 48.0]*** & $-$18.0 & [$-$27.8, $-$8.3]*** & $-$16.8 & [$-$26.4, $-$7.2]** & $-$3.6 & [$-$13.1, 5.8] \\
    \addlinespace[3pt]
    \midrule
    Non-OE &Accuracy & 22.6 & [15.4, 29.7]*** & $-$16.2 & [$-$25.6, $-$6.6]*** & $-$3.6 & [$-$12.5, 5.4] & $-$4.6 & [$-$13.9, 4.4] \\
    users only & Clinical util. & 28.7 & [18.8, 38.1]*** & $-$18.6 & [$-$31.3, $-$5.7]** & $-$15.5 & [$-$27.2, $-$3.5]* & 3.9 & [$-$8.7, 16.5] \\
    & Source Qual. & 44.3 & [36.2, 52.0]*** & $-$23.8 & [$-$35.3, $-$12.0]*** & $-$22.3 & [$-$33.3, $-$11.4]*** & $-$0.9 & [$-$12.5, 10.6] \\
    & Completeness & 30.6 & [21.1, 39.9]*** & $-$26.3 & [$-$38.2, $-$14.4]*** & $-$0.5 & [$-$12.9, 11.5] & $-$6.4 & [$-$19.1, 5.8] \\
    & Verifiability & 33.7 & [26.2, 41.2]*** & $-$8.6 & [$-$19.0, 1.9] & $-$17.7 & [$-$27.5, $-$7.9]*** & $-$9.8 & [$-$19.4, $-$0.4] \\
    \addlinespace[3pt]
    \midrule
    Sensitivity: & Accuracy & 24.6 & [18.4, 30.7]*** & $-$22.7 & [$-$30.4, $-$15.0]*** & $-$7.5 & [$-$14.9, $-$0.3]* & $-$4.8 & [$-$12.1, 2.5] \\
    bootstrap & Clinical util. & 29.9 & [21.9, 37.7]*** & $-$23.3 & [$-$33.9, $-$12.9]*** & $-$19.8 & [$-$29.1, $-$10.5]*** & 0.8 & [$-$9.6, 11.5] \\
    by grader & Source Qual. & 38.7 & [31.5, 45.6]*** & $-$26.8 & [$-$36.6, $-$17.4]*** & $-$19.8 & [$-$28.8, $-$11.0]*** & $-$7.9 & [$-$17.1, 0.9] \\
    & Completeness & 30.8 & [22.9, 38.4]*** & $-$32.2 & [$-$41.2, $-$22.9]*** & $-$10.3 & [$-$19.3, $-$1.2]* & $-$1.4 & [$-$11.3, 8.3] \\
    & Verifiability & 26.3 & [19.6, 32.8]*** & $-$21.2 & [$-$29.6, $-$13.0]*** & $-$14.0 & [$-$21.4, $-$6.4]*** & $-$2.0 & [$-$9.5, 5.5] \\
    \bottomrule
  \end{tabular}
  \begin{flushleft}
  \scriptsize OE = OpenEvidence; Claude = Claude Opus 4.8; Gemini = Gemini 3.1 Pro.
  \end{flushleft}
  \label{tab:p_values}
\end{table}

\begin{landscape}
\begin{figure}[p]
  \centering
  \includegraphics[width=0.98\linewidth]{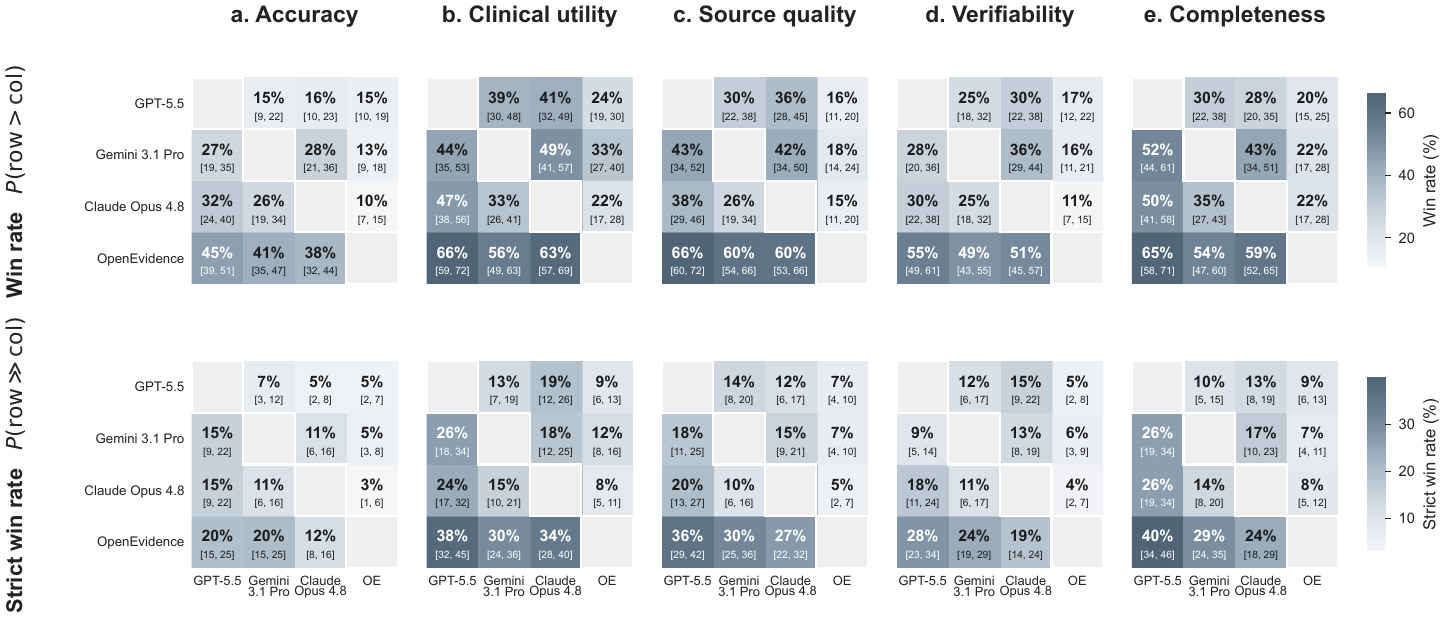}
  \caption{
  \textbf{Pairwise head-to-head win rates for all five evaluation axes.} Each column reports the pairwise win rate $P(\mathrm{row} \succ \mathrm{column})$ for every pair of systems on one axis: \textbf{a}, accuracy; \textbf{b}, clinical utility; \textbf{c}, source quality; \textbf{d}, verifiability; \textbf{e}, completeness.
  The top row defines wins as the row system being rated somewhat or strongly better than the column system, while the bottom row requires strict wins where the row system must be strongly preferred.
  Cells are additionally annotated with a bootstrap 95\% CI; darker cells denote higher win rates, and the diagonal (self-comparison) is omitted. Systems are GPT-5.5, Gemini 3.1 Pro, Claude Opus 4.8 and OpenEvidence (OE).}
  \label{fig:appendix_pairwise_matrices}
\end{figure}
\end{landscape}

\begin{figure}[tbp]
  \centering
  \includegraphics[width=0.9\linewidth]{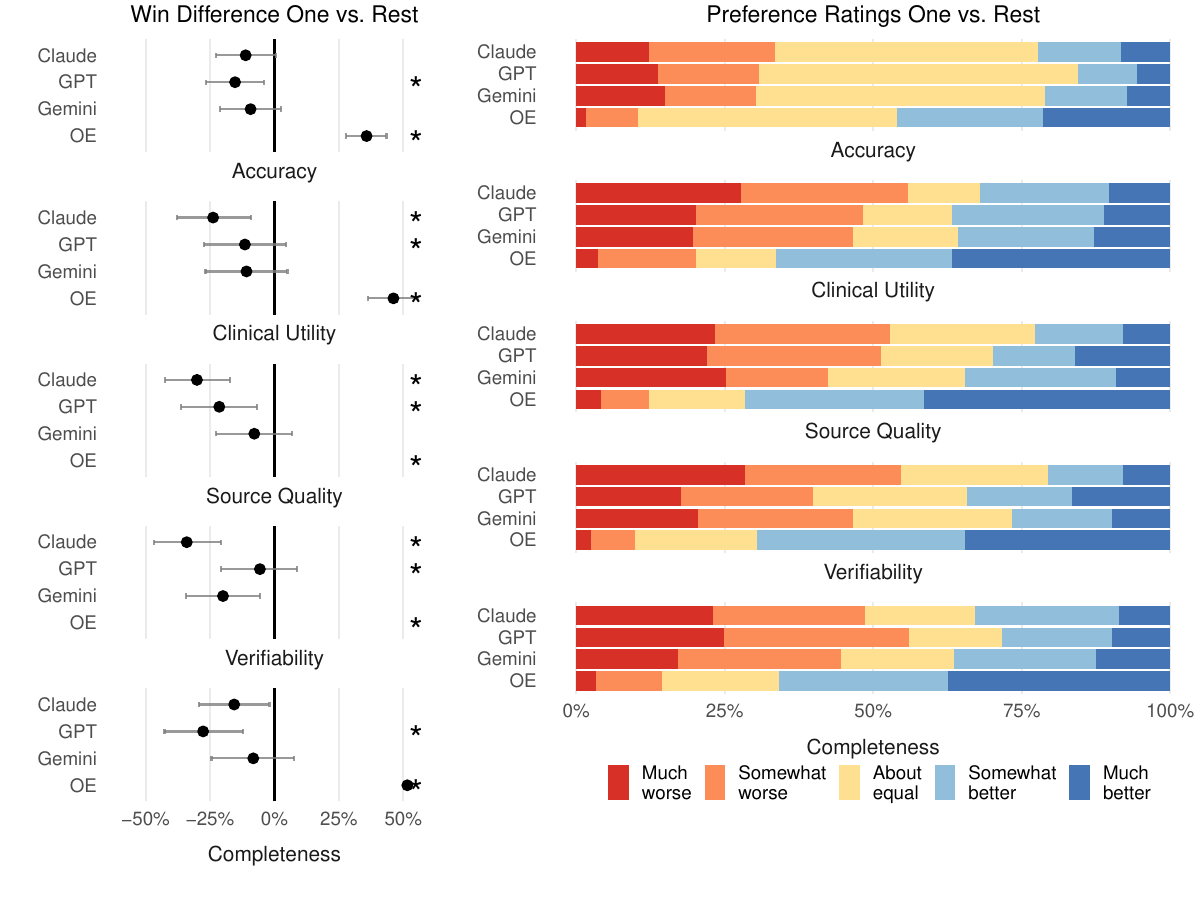}
  \caption{
  \textbf{One vs. Rest win differences  and preference rating distributions when AI systems provide supporting citations.}
  As a secondary analysis, AI systems were compared when their answers were rendered in the mode with text \textit{and} citations.
  Asterisks mark systems whose one-vs-rest win difference differs significantly from zero ($p<0.05$).
  }
  \label{fig:winrate_by_model_textcitations}
\end{figure}

\begin{figure}[tbp]
  \centering
  \includegraphics[width=0.9\linewidth]{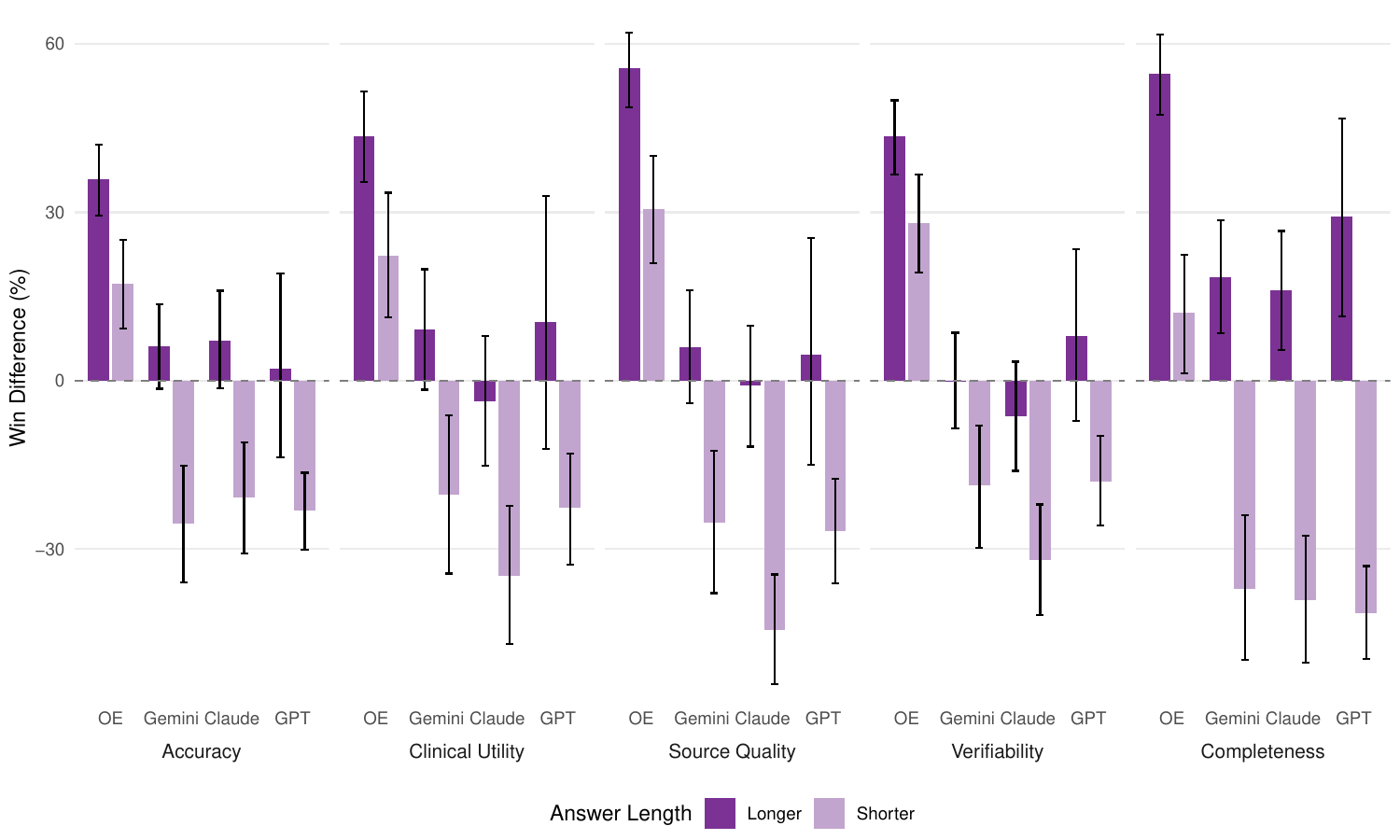}
  \includegraphics[width=0.9\linewidth]{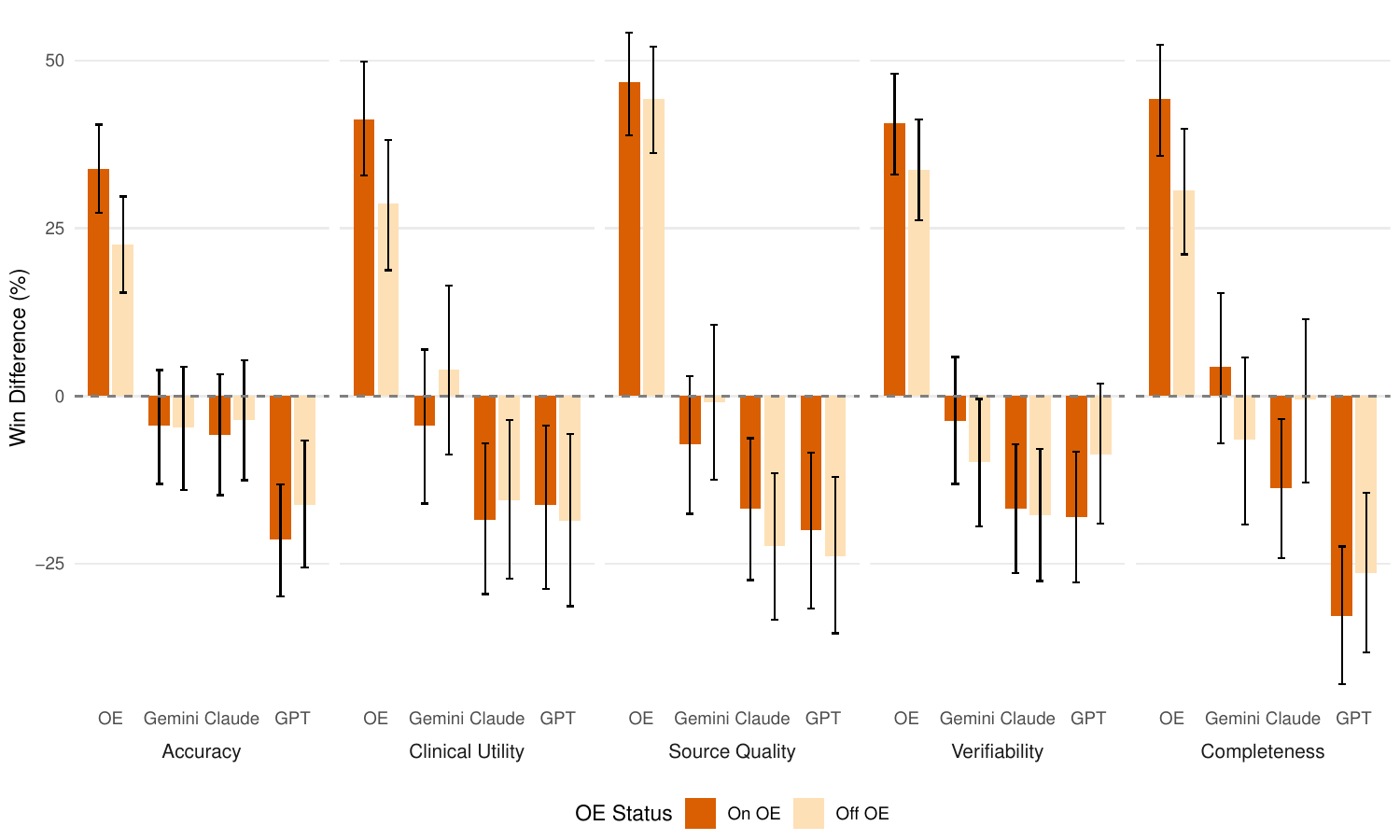}
  \caption{\textbf{One-vs-rest win difference by evaluation axis and AI system, stratified by answer length (top) and evaluator OpenEvidence user status (bottom).}
  For a given system A, the one-vs-rest win difference is its percentage of wins minus losses across head-to-head comparisons against other systems (the dashed line marks zero, where wins and losses are balanced). \textbf{Top}, comparisons stratified by whether system A's answer was longer (dark) or shorter (light) than its comparator. \textbf{Bottom}, comparisons stratified by whether the evaluating clinician was a registered OpenEvidence user (On OE, dark) or not (Off OE, light). Bars are grouped by evaluation axis (accuracy, clinical utility, source quality, verifiability and completeness) for GPT-5.5, Gemini 3.1 Pro, Claude Opus 4.8 and OpenEvidence; error bars are 95\% question-clustered bootstrap confidence intervals. Results pool the two rendering modes.}
  \label{fig:length_stratified_winrate_by_model}
\end{figure}

\begin{figure}[tbp]
  \centering
  \includegraphics[width=0.9\linewidth]{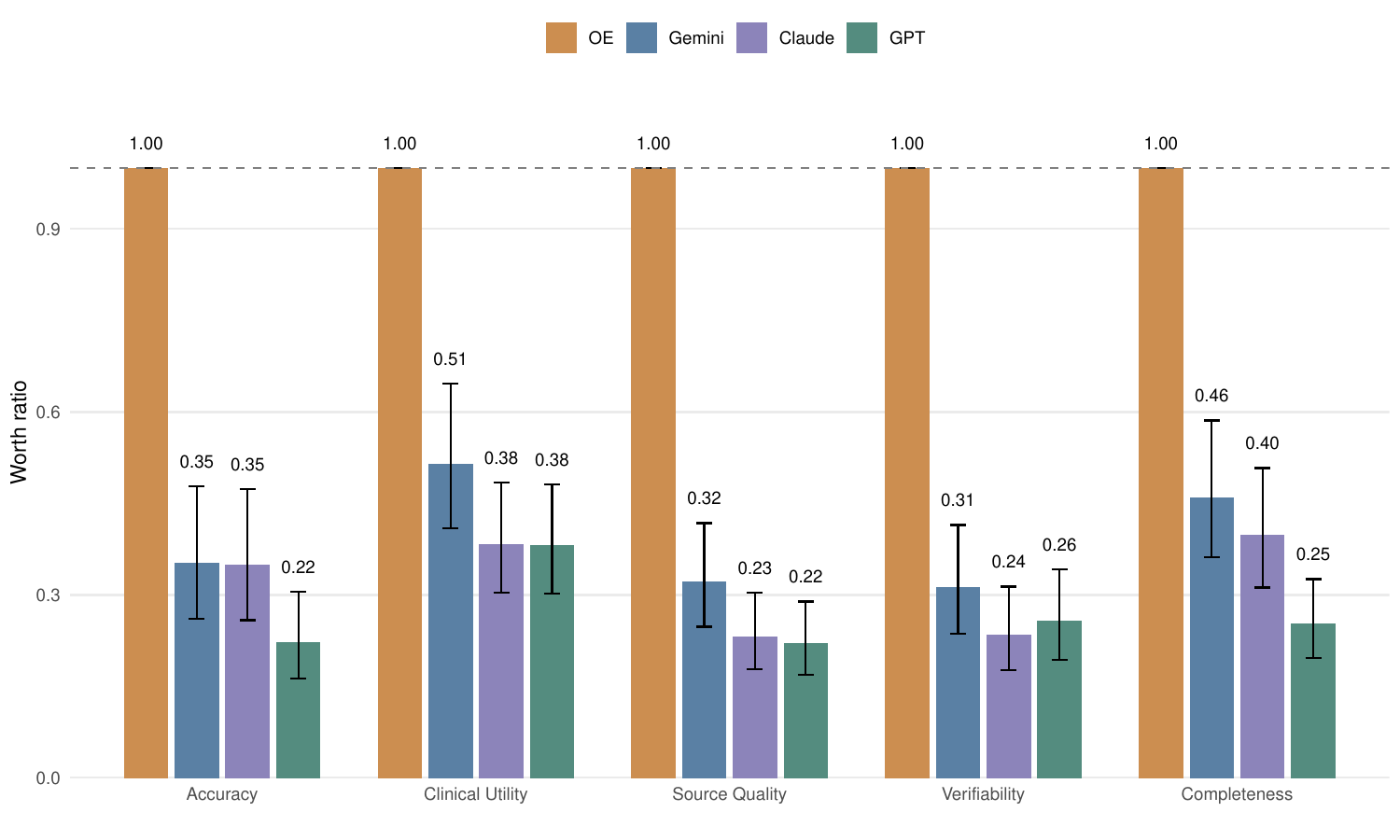}
  \caption{
  \textbf{Davidson tie-model worth ratios by evaluation axis and AI system.} Worth ratios from the Davidson tie model \citep{Davidson1970-hf} (extension of Bradley-Terry \citep{Bradley1952-kj}), fit separately per axis.
  Bars show each system's worth ratio per the Davidson model, normalised to the top-ranked system on that axis (OpenEvidence (OE)), which is fixed at the reference value of 1.00 (dashed line); a value of, e.g., 0.40 denotes 40\% of OE's estimated worth. Error bars are 95\% Wald confidence intervals; no interval is shown for the reference system.}
  \label{fig:davidson_worth_by_model}
\end{figure}

\pagebreak
\section*{Acknowledgements}
The authors thank OpenEvidence for sharing their data for this study and for sharing it with the broader research community.

\section*{Author Contributions}
JF, YM, AJ, and VS designed the study and data collection plan.
JF, YM, PH, PV, and JO designed the statistical analysis plan and implemented/verified of the analysis code.
JF conducted analyses blinded.
JF and VP drafted the manuscript.
All authors contributed with interpreting analyses and manuscript editing.

\pagebreak

\bibliographystyle{unsrtnat}
\bibliography{paperpile}

\end{appendices}

\end{document}